 \documentclass[nohyperref]{article}

 \usepackage{microtype}
 \usepackage{graphicx}
 \usepackage{subfigure}
 \usepackage{booktabs} 
 \usepackage{xspace}
 \usepackage{hyperref}

 \usepackage[accepted]{icml2023}
 
 \usepackage{amsmath}
 \usepackage{amssymb}
 \usepackage{mathtools}
 \usepackage{amsthm}
 
 \usepackage[capitalize,noabbrev]{cleveref}
 
 \theoremstyle{plain}

 \theoremstyle{definition}

 \theoremstyle{remark}

 \usepackage[textsize=tiny]{todonotes}

 \newcommand{\opt}{\mathcal{O}}

 \newcommand{\bx}{\mathbf{x}}
 
 \newcommand{\clarecomment}[1]{}
 \newcommand{\willcomment}[1]{}
 \newcommand{\zeyucomment}[1]{}
 
 \newcommand{\states}{\mathcal{S}}
 \newcommand{\actions}{\mathcal{A}}
 \icmltitlerunning{Understanding Plasticity in Neural Networks}

\begin{document}
 
 \twocolumn[
 \icmltitle{Understanding Plasticity in Neural Networks}
 
 
 
 \begin{icmlauthorlist}
 \icmlauthor{Clare Lyle}{dm}
 \icmlauthor{Zeyu Zheng}{dm}
 \icmlauthor{Evgenii Nikishin}{mila}
 
 \icmlauthor{Bernardo Avila Pires}{dm}
 \icmlauthor{Razvan Pascanu}{dm}
 \icmlauthor{Will Dabney}{dm}
 \end{icmlauthorlist}
 
 \icmlaffiliation{dm}{Google DeepMind}
 \icmlaffiliation{mila}{Montreal Institute for Learning Algorithms (work done during internship at DeepMind)}
 \icmlcorrespondingauthor{Clare Lyle}{clarelyle@deepmind.com}
 \icmlkeywords{Machine Learning, ICML}
 
 \vskip 0.3in
 ]
 
 
 
 \printAffiliationsAndNotice{} 
 
 \begin{abstract}
 Plasticity, the ability of a neural network to quickly change its predictions in response to new information, is essential for the adaptability and robustness of deep reinforcement learning systems. Deep neural networks are known to lose plasticity over the course of training even in relatively simple learning problems, but the mechanisms driving this phenomenon are still poorly understood. This paper conducts a systematic empirical analysis into plasticity loss, with the goal of understanding the phenomenon mechanistically in order to guide the future development of targeted solutions. We find that loss of plasticity is deeply connected to changes in the curvature of the loss landscape, but that it often occurs in the absence of saturated units. Based on this insight, we identify a number of parameterization and optimization design choices which enable networks to better preserve plasticity over the course of training. We validate the utility of these findings on larger-scale RL benchmarks in the Arcade Learning Environment.
 \end{abstract}
 
 \section{Introduction}
 
 It is a widely observed phenomenon that after training on a non-stationary objective, neural networks exhibit a reduced ability to solve new tasks \citep{lyle2021understanding, nikishin2022primacy, dohare2021continual}.
 This loss of plasticity occurs most robustly when the relationship between inputs and prediction targets changes over time, and the network must learn to `overwrite' its prior predictions \citep{lyle2021understanding}. While such scenarios are relatively rare in supervised learning, they are baked into the way that deep reinforcement learning (RL) agents are trained. Understanding how plasticity is lost, and whether this loss can be mitigated, is crucial if we wish to develop deep RL agents which can rise to the challenge of complex and constantly-changing environments.
 Existing methods to promote trainability act on a wide variety of potential mechanisms by which plasticity might be lost, including resetting of layers \citep{nikishin2022primacy} and activation units \citep{dohare2021continual}, and regularization of the features \citep{kumar2020implicit, lyle2021understanding}. 
 While all of these works observe performance improvements, it is unlikely that they are all obtaining these improvements by the same mechanism. As a result, it is difficult to know how to improve on these interventions to further preserve plasticity. 
 
 This paper seeks to identify the mechanisms by which plasticity loss occurs. We begin with an analysis of two interpretable case studies, illustrating the mechanisms by which both adaptive optimizers and naive gradient descent can drive the loss of plasticity. Prior works have conjectured, implicitly or explicitly, that a variety of network properties might cause plasticity loss: we present a falsification framework inspired by the study of causally robust predictors of generalization \citep{dziugaite2020search}, and leverage this framework to show that loss of plasticity cannot be uniquely attributed to any of these properties.
 While difficult to characterize explicitly, we provide evidence that the curvature of the loss landscape induced by new tasks on trained parameters is a crucial factor determining a network's plasticity, particularly in value-based reinforcement learning algorithms.
 
 We conclude by completing a broad empirical analysis of methods which aim to improve the ability of a network to navigate the loss landscape throughout training. We find that architectural choices which have been conjectured to smooth out the loss landscape, such as categorical output representations and normalization layers, provide the greatest improvements to plasticity, while methods which perturb the parameters or provide other forms of regularization tend to see less benefit. To test the generality of these findings, we apply the best-performing intervention, layer normalization, to a standard DQN architecture and obtain significant improvements in performance on the Arcade Learning Environment benchmark. We conclude that controlling the loss landscape sharpness and optimizer stability present highly promising avenues to improve the robustness and usability of deep RL methods. 
 
 \section{Background}
 
 It has long been observed that training a network first on one task and then a second will result in reduced performance on the first task \citep{french1999catastrophic}. This phenomenon, known as catastrophic forgetting, has been widely studied by many works \citep{kirkpatrick2017overcoming, lee2017overcoming, kemker2018measuring}. This paper concerns itself with a different phenomenon: in certain situations, training a neural network on a series of distinct tasks can result in worse performance on later tasks than what would be obtained by training a randomly initialized network of the same architecture.
 
 \subsection{Preliminaries}
 
 \textbf{Temporal difference learning.} Plasticity loss naturally arises under non-stationarity; we will focus our analysis on temporal difference (TD) learning with neural networks, a setting known to induce significant non-stationarity. We assume the standard reinforcement learning problem of an agent interacting with an environment $\mathcal{M}$, with observation space $\mathcal{S}$, action space $\mathcal{A}$, reward $R$ and discount factor $\gamma$, with the objective of maximizing cumulative discounted reward \citep{sutton2018reinforcement}. Networks trained via temporal difference learning receive as input sampled \textit{transitions} from an agent's interaction with the environment, of the form $\tau_t = (s_{t-1}, a_{t}, r_t, s_{t})$, where $s_{t-1}, s_{t} \in \states$, $a_t \in \actions$, and $r_t = R(s_t)$. We let $\theta'$ denote the \textit{target parameters}; in practice, $\theta'$ is usually an outdated copy of $\theta$ from a previous iteration, but other choices include setting it to be equal to the current parameters, or using a moving average of past values. The network $f:\Theta \times \states \times \actions \rightarrow \mathbb{R}$ is trained to minimize the temporal difference error
 \begin{equation}\label{eq:td}
     \ell(\theta, \tau_t) = \bigg \|  f(\theta, s_{t-1}, a_t) - \square (r_t + \gamma f(\theta', s_t, a')) \bigg \|^2 
 \end{equation} 
 where $\square$ denotes a stop-gradient, $\gamma<1$ is the discount factor, and $a'$ is chosen based on the variant of TD learning used. Crucially, the regression target ${r_t + \gamma  f(\theta', s_t, a')}$ depends on the parameters $\theta'$ and changes as learning progresses. This nonstationarity occurs even if the policy and input distribution are fixed, meaning that we can study the role of nonstationarity independent of the agent's exploration strategy. We will use the shorthand $\ell(\theta)$, with $\theta'$ implicit, for the expectation of this loss over some input distribution. 
 
 \textbf{Loss landscape analysis.} We will be particularly interested in the study of the structure of the loss landscape traversed by an optimization algorithm. We will leverage two principal quantities in this analysis: the Hessian of the network with respect to some loss function, and the gradient covariance. The Hessian of a network $f$ at parameters $\theta$ with respect to some loss $\ell(\theta)$ is the matrix defined as 
 \begin{equation}
     H_\ell (\theta) = \nabla^2_\theta \ell(\theta) \in \mathbb{R}^{d \times d}
 \end{equation}
 where $d=|\theta|$ is the number of parameters. Of particular relevance to optimization is the eigenspectrum of the Hessian $\Lambda(H_\ell(\theta) ) = (\lambda_1 \ge \dots \ge \lambda_d)$. The maximal eigenvalue, $\lambda_1$, can be interpreted as measuring the sharpness of the loss landscape \citep{dinh2017sharp}, and the condition number $\lambda_1/\lambda_d$ has significant implications for convergence of gradient descent optimization in deep neural networks \citep{gilmer2022loss}.
 
 We will also take interest in the covariance structure of the gradients of different data points in the input distribution, a property relevant to both optimization and generalization  \citep{fort2019stiffness,lyle2022learning}. We will estimate this covariance structure by sampling $k$ training points $\bx_1, \dots, \bx_k$, and computing the matrix $C_k \in \mathbb{R}^{k \times k}$ defined entrywise as
 
 \begin{equation}C_k[i, j] = \frac{\langle \nabla_\theta \ell(\theta, \bx_i), \nabla_\theta \ell(\theta, \bx_j) \rangle}{\|\nabla_\theta \ell(\theta, \bx_i)\| \|\nabla_\theta \ell(\theta, \bx_j)\|} \; . \end{equation}
 
 If the off-diagonal entries of $C_k$ contain many negative values, this indicates interference between inputs, wherein the network cannot reduce its loss on one subset without increasing its loss on another. If the matrix $C_k$ exhibits low rank (which, given a suitable ordering $\sigma$ of the data points $\bx_{\sigma(1)}, \dots, \bx_{\sigma(k)}$ will yield a block structure) then the gradients are approximately colinear, which can indicate either generalization when their dot product is positive, or interference when their dot product is negative.
 
 \subsection{Defining plasticity}
 \label{sec:plasticity-definition}
 Plasticity, broadly construed, refers to a neural network's ability to learn new things.  The study of plasticity has concerned neuroscience for several decades \citep{mermillod2013stability, abbott2000synaptic}, but has only recently emerged as a topic of interest in deep learning \citep{berariu2021study, ash2020warm, delfosse2021adaptive, sodhani2020toward}. Classical notions of complexity from the computational learning theory literature \citep{vapnik1968uniform, bartlett2002rademacher} evaluate whether a hypothesis class contains functions that capture arbitrary patterns, but are agnostic to the ability of a particular search algorithm, such as gradient descent, to find them, making them unsuitable proxies for plasticity. A billion-parameter neural network architecture might have the \textit{capacity} to represent a rich class of functions, but if all of its activation units are saturated then it cannot be trained by gradient descent to realize this capacity.
 
 Studies of plasticity in both supervised and reinforcement learning have observed reduced generalization performance as a result of overfitting to limited data early in training \citep{ash2020warm, berariu2021study, igl2021transient}. Many works have further identified an impaired ability to even reduce the learning objective on the training distribution \citep{dohare2021continual, lyle2021understanding, nikishin2022primacy}. This work will leverage the formulation of \citet{lyle2021understanding}, who define plasticity as the ability of a network to update its predictions in response to a wide array of possible learning signals on the input distribution it has been trained on. This formulation is applicable to learning problems which do not admit a straightforward train-test split, as is the case in many deep RL environments. 
 
 Concretely, we consider an optimization algorithm ${\opt : (\theta, \ell) \mapsto \theta^*}$ which takes initial parameters $\theta \in \Theta$ and some objective function $\ell: \Theta \rightarrow \mathbb{R}$, and outputs a new set of parameters $\theta^*$. The parameters $\theta^*$ need not be an optimum: $\mathcal{O}$ could, for example, run gradient descent for five steps. In order to measure the flexibility with which a network can update its predictions under this optimization algorithm, we consider a distribution over a set of loss functions $\mathcal{L}$ each defined by some learning objective. For example, consider a distribution over regression losses 
 \begin{equation}\label{eq:plasticity}
     {\ell_{f, \mathbf{X}}(\theta) = \mathbb{E}_{x \sim \mathbf{X}} [(f(\theta, \bx) - g_{\omega}(\bx))^2]}
 \end{equation}
 where $g_\omega$ is induced by a random initialization $\omega$ of a neural network. In order to match the intuition that more adaptable networks should have greater plasticity, we set a baseline value $b$ to be the loss obtained by some baseline function (e.g. if $\ell$ is a regression loss on some set of targets, we set $b$ to be the variance of the targets), and then define plasticity to be the difference between the baseline and the expectation of the final loss obtained by this optimization process after starting from an initial parameter value $\theta_t$ and optimizing a sampled loss function $\ell$ subtracted from the baseline $b$. 
 \begin{equation}
     \mathcal{P}(\theta_t) =  b - \mathbb{E}_{\ell \sim \mathcal{L}}[ \ell(\theta^*_t) ] \text{ where } \theta_t^* = \mathcal{O}(\theta_t, \ell)
 \end{equation}
 We then define the loss of plasticity over the course of a trajectory $(\theta_t)_{t=0}^N$ as the difference $\mathcal{P}(\theta_t) - \mathcal{P}(\theta_0)$. We note that this definition of plasticity loss is independent of the value of the baseline $b$, i.e. the difficulty of the probe task for the network, allowing us to measure the relative change in performance of checkpoints taken from a training trajectory.
 
 \section{Methodology and Motivating Questions}
 
 The following sections will present a series of experiments which tease apart different causal pathways by which plasticity loss occurs and evaluate the predictive power of a range of hypotheses concerning the root causes thereof. We now outline the experimental methodology and research questions underpinning this investigation.
 
 \subsection{Measuring plasticity}
 \label{sec:measuring-plasticity}
 In order to determine whether a candidate intervention preserves plasticity, we must first set out a consistent standard by which we will measure plasticity. Given a suitably generic class of target functions inducing the losses $\ell$, ~\eqref{eq:plasticity} characterizes the adaptability of the network to arbitrary new learning signals from the unknown set of possible future tasks. We therefore construct a distribution over regression targets which corresponds to this uniform prior over possible future update directions. A different distribution over future target functions might give different numerical results; however, we believe that a uniform distribution captures a more universal notion of plasticity. 
 
 In our empirical evaluations, we will set $\mathbf{X}$ to be the set of transitions gathered by an RL agent and stored in some replay buffer, and $f$ to be a neural network architecture. Given some offset $a \in \mathbb{R}$, we will apply the transformation $g(x) = a + \sin(10^5 f(\bx; \omega_0))$, with $\omega_0$ sampled from the same distribution as $\theta_0$, to construct a challenging prediction objective which measures the ability of the network to perturb its predictions in random directions sampled effectively uniformly over the input space. Because the mean prediction output by a deep RL network tends to evolve away from zero over time as the policy improves and the reward propagates through the value function, we will set $a$ to be equal to the network's mean prediction in order not to bias the objective in favour of random initializations, which have mean much closer to zero. The optimizer $\mathcal{O}$ will be identical to that used by the network on its primary learning objective, and we found that running this optimizer for a budget of two thousand steps minimized iteration time while also providing enough opportunity for networks starting from random initializations to solve the task.
 \subsection{Environments}
 \label{sec:classification-mdps}
 We construct a simple MDP analogue of image classification, i.e. the underlying transition dynamics are defined over a set of ten states and ten actions, and the reward and transition dynamics depend on whether or not the action taken by the agent is equal to the index of its corresponding state. We construct three variants of a block MDP whose state space is the discrete set $\{0, \dots, 9\}$ and whose observation space is given by either the CIFAR-10 or MNIST dataset.
 
 \textbf{True-label:} each state $s$ of the MDP produces an observation from that class in the underlying classification dataset. Given action $a$, the reward is the indicator function $\delta_{a=s}$. The MDP then randomly transitions to a new state.
 
 \textbf{Random-label:} follows the same dynamics as the previous environment, but each image is assigned a random label in $\{0 \dots 9\}$, and the observation from an MDP state $i$ is sampled from images with (randomized) label $i$.
 
 \textbf{Sparse-reward:} exhibits the same observation mapping as \textit{true-label}. The reward is equal to $\delta_{a = s = 9}$. The MDP transitions to a random state if $a \neq s$ and otherwise to $s+1$.
 
 We design these environments to satisfy two principal desiderata: first, that they present visually interesting prediction challenges with varying degrees of reward smoothness and density, and second that they allow us to isolate non-stationarity due to policy and target network updates independent of a change in the state visitation distribution. 
 In the true-label and random-label variants, the transition dynamics do not depend on the agent's action, whereas in the sparse environment the policy influences the state visitation distribution. The different reward functions allow us to compare tasks which are aligned with the network's inductive bias (in the true-label task) and those which are not (the random-label task).
 
 \subsection{Outline of experimental results}
 
 The experiments presented in Sections~\ref{sec:case-studies} and ~\ref{sec:explanations} aim to answer a fundamental question:
 \textbf{what happens when neural networks lose plasticity?} Section~\ref{sec:case-studies} constructs two experimental settings which illuminate phenomena driving two very different forms of plasticity loss. The first constructs a non-stationary learning problem that induces extreme forms of instability in adaptive optimizers. The second illustrates a bias in the dynamics of gradient descent which leads to a progressive sharpening of the loss landscape of not just the current task, but also new tasks.
 
 Section~\ref{sec:explanations} asks \textbf{what properties \textit{cause} plasticity loss?} Disentangling cause from correlation is a notoriously difficult problem throughout science and economics. We put a number of quantities which have been conjectured to drive plasticity loss to the test, evaluating quantities such as weight norm, feature rank, and the number of dead units in the network on the tasks outlined in Section~\ref{sec:classification-mdps}. We follow up the largely negative results of these experiments with a qualitative analysis of learning curves on the probe tasks described in Section~\ref{sec:measuring-plasticity} that emphasizes the critical role of the loss landscape in plasticity. 
 
 Section~\ref{sec:interventions} addresses the question: \textbf{how can we mitigate plasticity loss?} It evaluates the effectiveness of a range of interventions on the network architecture and on the optimization protocol, focusing on methods known to increase the smoothness of the loss landscape, applying the same evaluation protocol as described in this section in order to measure plasticity across our classification MDP testbed. 
 
 \section{Two Simple Studies on Plasticity}
 \label{sec:case-studies}
 We begin with some interpretable examples of learning problems where plasticity loss occurs. These examples illustrate how the design of optimizers can interact with nonstationarity to destabilize training, and explore how the dynamics of gradient-based optimizers might affect more subtle properties of the loss landscape.
 \begin{figure}
     \centering
     \includegraphics[width=0.98\linewidth]{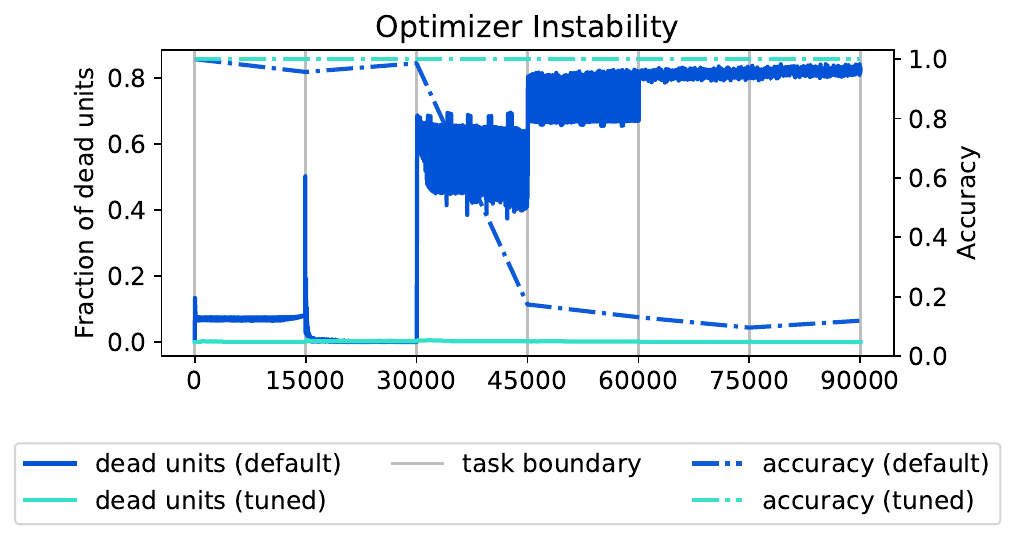}
     \caption{Abrupt task changes can drive instability in optimizers which depend on second-order moment estimates for adaptive learning rate scaling. Setting these estimators to be more robust to small gradient norms and to update moment estimates more quickly mitigates this issue.}
     \label{fig:Adam-sketch}
 \end{figure}
 
 \subsection{Optimizer instability and non-stationarity}
 
 The robustness of existing optimizers across a wide range of datasets and network architectures has played a key role in the widespread adoption of deep learning methods. For example, the Adam optimizer \citep{kingma2015Adam} with a learning rate of $10^{-3}$ will often yield reasonable initial results on a range of network architectures from which the practitioner can iterate. However, when the assumptions on stationarity underlying the design of this optimizer no longer hold, the optimization process can experience catastrophic divergence, killing off most of the network's ReLU units. We can see an example of this in a simple non-stationary task in Figure~\ref{fig:Adam-sketch}. A two-hidden-layer fully-connected neural network is trained to memorize random labels of MNIST images (full details provided in Appendix~\ref{sec:case-study-details}). After a fixed training budget, the labels are re-randomized, and the network continues training from its current parameters. This process quickly leads a default Adam optimizer to diverge, saturating most of its ReLU units and resulting in trivial performance on the task that a freshly 
 
 The mechanism of this phenomenon emerges when we consider the update rule for Adam, which tracks a second-order estimate $\hat{v}_t$ along with a first-order moment estimate $\hat{m}_t$ of the gradient via an exponential moving average 
 \begin{equation}
     \label{eq:Adam}
     u_t = \alpha \frac{\hat{m}_t}{\sqrt{\hat{v}_t + \bar{\epsilon}} + \epsilon}.
 \end{equation}
 
 Gradients tend to have norm proportional to the training loss. When the loss changes suddenly, as is the case when the perfectly-memorized MNIST labels are re-randomized (or when the target network is updated in an RL agent), $\hat{m}_t$ and $\hat{v}_t$ will no longer be accurate estimates of their moment distributions. Under the default hyperparameters for deep supervised learning, $\hat{m}_t$ is updated more aggressively than $\hat{v}_t$, and so the updates immediately after a task change will scale as a large number divided by a much smaller number, contributing to the instability we observe in Figure~\ref{fig:Adam-sketch}. In this instance, the solution is simple: we simply increase $\epsilon$ and set a more aggressive decay rate for the second-moment estimate, and the network avoids catastrophic instability. Intriguingly, a large value of $\epsilon$ is frequently used in deep RL algorithms such as DQN \citep{mnih2015human} relative to the default provided by optimization libraries, suggesting that the community has implicitly converged towards optimizer hyperparameters which promote stability under nonstationarity. 
 initialized network could solve perfectly.
 
 \subsection{Loss landscape evolution under non-stationarity}
 \begin{figure}
     \centering
     \includegraphics[width=\linewidth]{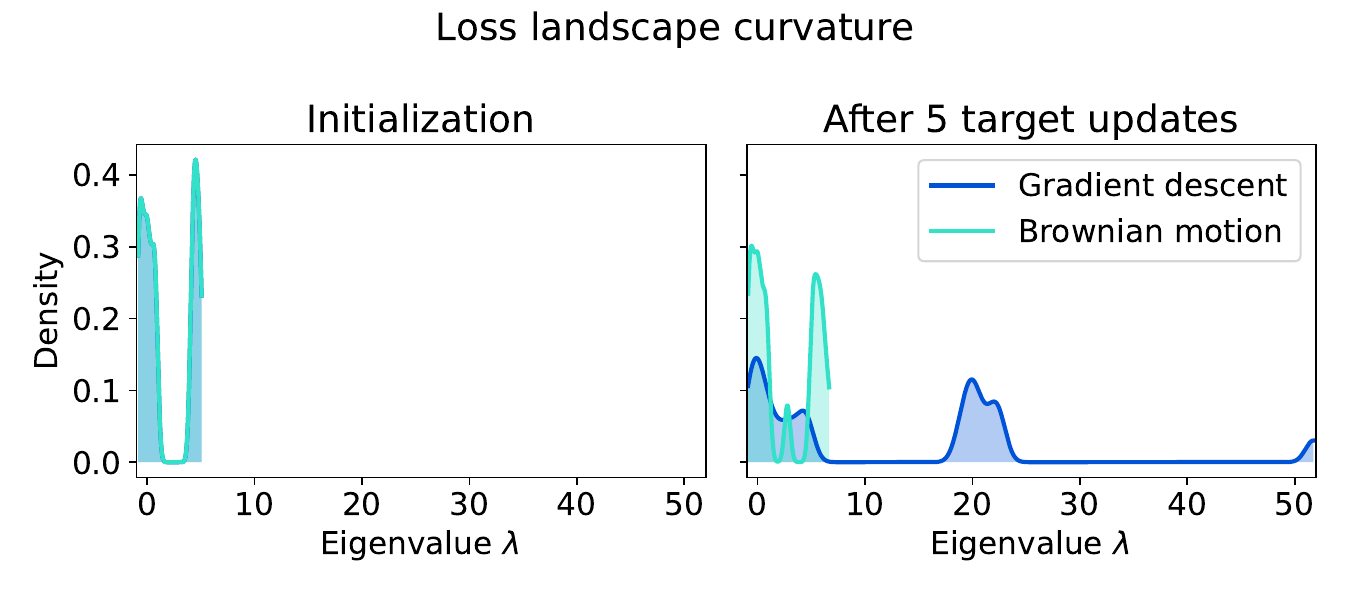}\\
     \includegraphics[width=1.12\linewidth]{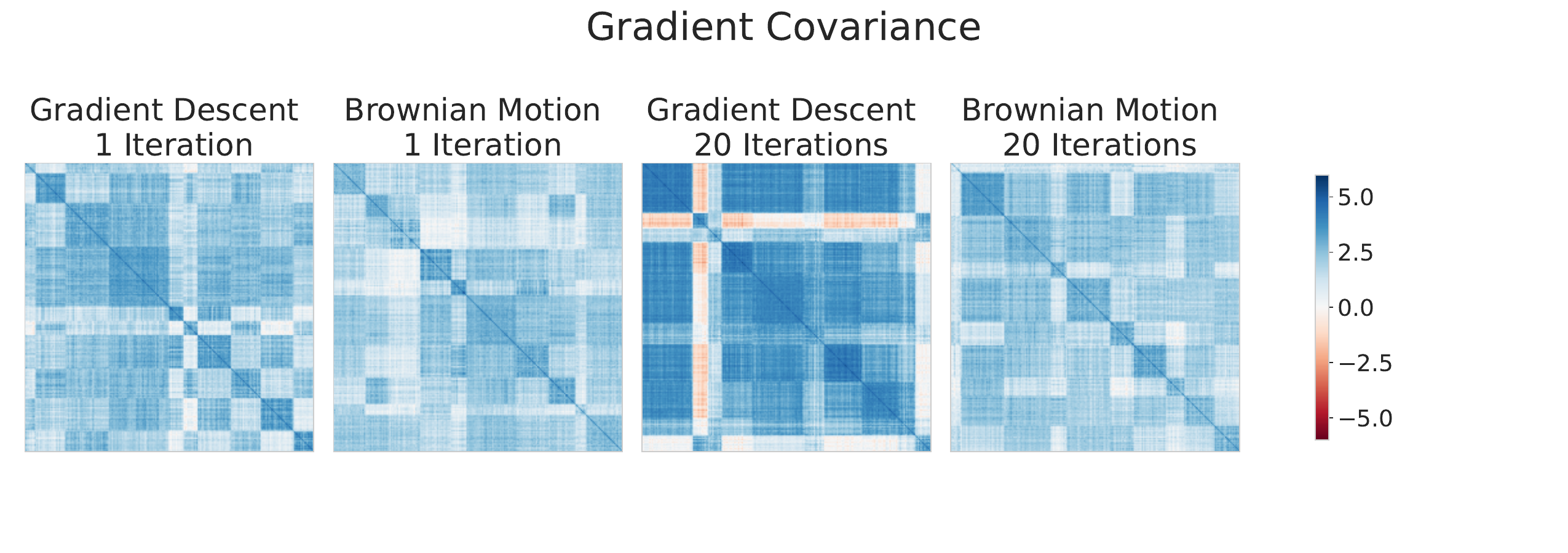}
     \caption{Evolution of the gradient and Hessian under gradient-based optimization compared to random perturbation of the parameters. Top: the density of the spectrum of the Hessian over different values of $\lambda$ exhibits a larger outlier peak after gradient descent. Bottom: gradient descent induces more gradient interference between inputs and greater curvature of the loss landscape.
     }
     \label{fig:hessian-random-walk}
 \end{figure}
 Even when optimization is sufficiently stable to avoid saturated units, prior work still observes reductions in network plasticity \citep{lyle2021understanding}. The causes of this phenomenon are more difficult to tease apart; neural network initializations have been tuned over several decades to maximize trainability, and many properties of the network change during optimization which could be driving the loss of plasticity. A natural question we can ask in RL is whether the optimization dynamics followed by a network bias the parameters to become less trainable, or whether the loss of plasticity is a natural consequence of any perturbation away from a carefully chosen initialization distribution.
 
 We frame this question as a controlled experiment, in which we compare the evolution of two coupled updating procedures: one follows gradient-based optimization on a non-stationary objective (full details in Appendix~\ref{sec:case-study-details}); the second follows a random walk, where we add a Gaussian perturbation to the parameters with norm equal to the size of the gradient-based optimizer update. Both trajectories start from the same set of randomly initialized parameters and apply updates of equal norm; the only difference is the direction each step takes. We evaluate how the structure of the local loss landscape with respect to a probe task evolves in both networks by comparing the Hessian eigenvalue distribution, and by comparing the covariance structure $C_k$ of gradients on sampled inputs, with $k=512$ equal to the batch size used for training. We compute the Hessian matrix for a regression loss towards a perturbation $\epsilon \sim \mathcal{N}(0, 1)$ of the network's current output, i.e. ${\ell(\theta) = [f_\theta(\mathbf{X}) - \square f_{\theta}(\mathbf{X}) + \epsilon]^2}$ where $\square$ indicates a stop-gradient, to obtain a proxy for how easily the network can update its predictions in arbitrary directions; we do not evaluate the Hessian or gradient structure of the primary learning objective as these will trivially differ between the trajectories.
 
 We observe that the spectral norm of the Hessian of both processes increases over time; however, the outliers of the spectrum grow significantly faster in the network trained with gradient descent. Additionally, the network trained with gradient descent begins to exhibit negative interference between gradients, a phenomenon not observed in the Brownian motion. In other words, the inductive bias induced by gradient descent can push the parameters towards regions of the parameter space where the local loss landscape is less friendly to optimization towards arbitrary new objectives than what would be obtained by blindly perturbing randomly initialized parameters.
 
 \section{Explaining Plasticity Loss}
 \label{sec:explanations}
 While in some instances it is straightforward to deduce the cause of plasticity loss, most learning problems induce complex learning dynamics that make it difficult to determine root causes. This section will show that a number of plausible explanations of plasticity loss, including the rank of the network's features, the number of saturated units, the norm of its parameters, and the rank of the weight matrices, do not identify robust causal relationships. We provide some evidence supporting the hypothesis that plasticity loss arises due to changes in the network's loss landscape, and conclude with a discussion of the potential trade-offs that must be faced between preserving a trainable gradient structure and accurately predicting a value function.
 \begin{figure*}
     \centering
     \includegraphics[width=\linewidth]{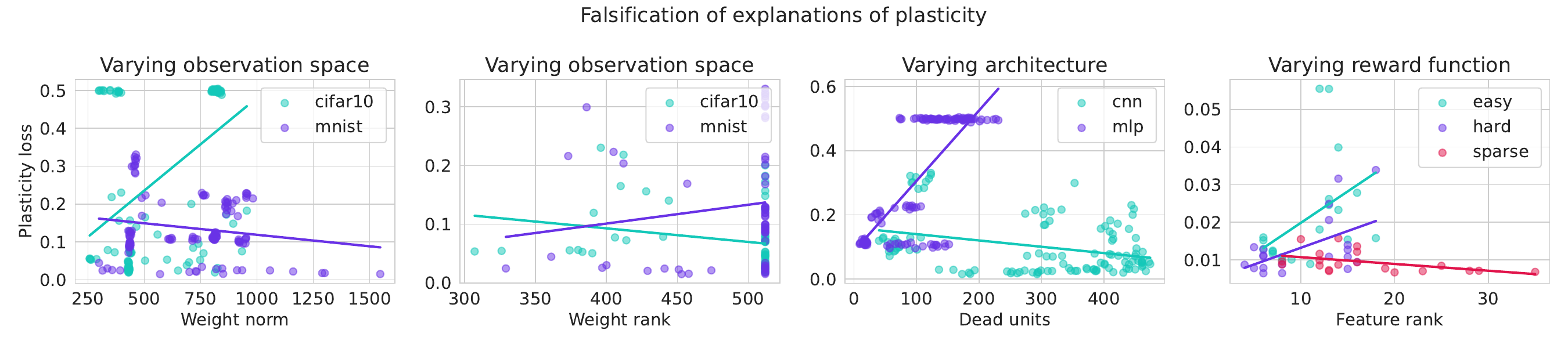} \vspace{-1em}
     \caption{Results of our experimental falsification design: for any variable we consider, it is possible to construct a set of learning problems in which the variable exhibits either a positive or a negative correlation with plasticity. For example, weight norm and weight rank exhibit differing correlation signs depending on the observation space, while feature rank and sparsity depend on the reward structure of the environment.}
     \label{fig:disproving}
 \end{figure*}

 \subsection{Experimental setting}
 \label{subsec:experimental_setting}
 
 We train a set of DQN agents on each environment-observation space combination in the classification MDP set described in Section~\ref{sec:classification-mdps}, and evaluate the ability of each network to fit a randomly generated set of target functions as described in Section~\ref{sec:plasticity-definition} after a fixed number of training steps. In the experiments shown here, we run the DQN agents with a target network update period of 1,000 steps; as mentioned previously, the changing bootstrap target is the principal source of non-stationarity in the true-label and random-label tasks. Every 5000 steps, we pause training, and from a copy of the current parameters $\theta_t$ we train the network on a set of new regression problems to probe its plasticity. We log the loss at the end of 2,000 steps of optimization, sampling 10 different random functions, then resume training of the RL task from the saved parameters $\theta_t$. We consider two network architectures: a fully-connected network (MLP) and a convolutional network architecture (CNN). Full details of the environments are included in Appendix~\ref{sec:toy-details}.

 \subsection{Falsification of prior hypotheses}
 Prior work has proposed a number of plausible explanations of why neural networks may exhibit reduced ability to fit new targets over time. Increased weight norm \citep{nikishin2022primacy}, low rank of the features or weights \citep{kumar2020implicit, gulcehreempirical}, and inactive features \citep{lyle2021understanding, dohare2021continual} have all been discussed as plausible mechanisms by which plasticity loss may occur. However, the explanatory power of these hypotheses has not been rigorously tested. While a correlation between a particular variable and plasticity loss can be useful for diagnosis, only a causal relationship indicates that intervening on that variable will necessarily increase plasticity.
 
 
 This section will seek to answer whether the above candidate explanations capture causal pathways. Our analysis is based on a simple premise: that for a quantity to exhibit explanatory power over plasticity loss, it should exhibit a consistent correlation across different experimental interventions \citep{buhlmann2020invariance}. If, for example, parameter norm is positively correlated with plasticity in one observation space and negatively correlated in another, then it can be ruled out as a causal factor in plasticity loss. 
 To construct this experiment, we train 128\clarecomment{?} DQN agents under a range of tasks, observation spaces, optimizers, and seeds. \clarecomment{TODO: go to xm and confirm experiment details.} Over the course of training, we log several statistics of the parameters and activations, along with the plasticity of the parameters at each logging iteration. 
 
 In Figure~\ref{fig:disproving}, we show scatterplots illustrating the relationship between plasticity and each statistic, where each point in the scatterplot corresponds to a single training run. We see that for each of four quantities, there exists a learning problem where the quantity positively correlates with plasticity, and one in which it exhibits a negative correlation. In many learning problems the correlation between plasticity loss and the quantity of interest is nonexistent. In all cases we note that the correlation with plasticity is already quite weak; even so, the ability to reverse the sign of this correlation is a further mark against the utility of these simple statistics as causal explanations of plasticity. For example, we see a positive correlation between weight norm and plasticity loss in environments which use CIFAR-10 observations, but a slight negative correlation in environments which sample observations from MNIST. Analogous reversals happen for the other variable considered. 
 
 \subsection{Loss landscape evolution during training}
 \begin{figure}
     \centering
     \includegraphics[width=0.98\linewidth]{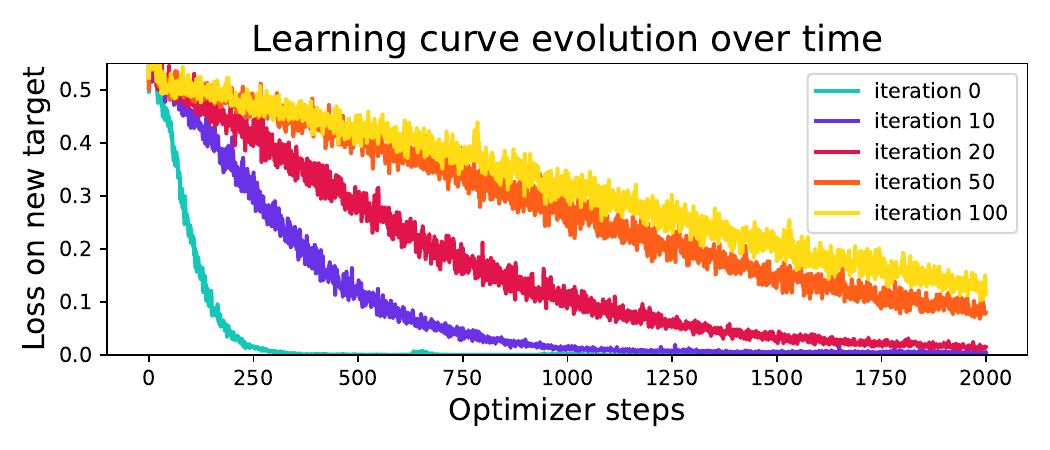}
     \caption{Plasticity loss corresponds to slower training progress, rather than higher plateaus, in the networks studied in this paper. We plot learning curves on a new target fitting task starting from network checkpoints at different points in training. This figure illustrates a CNN trained on the true-label MDP described in Section~\ref{sec:experiments} with a CIFAR-10 observation space.}
     \label{fig:learning_curves}
 \end{figure}
 
 If the simple statistics we have considered thus far lack explanatory power, how should we characterize plasticity loss? One open question is whether the reduced ability to fit arbitrary new targets arises because the optimization process gets caught in local optima, or whether it arises due to overall slow or inconsistent optimization progress. To answer this question, we turn our attention towards the learning curves generated by networks as they train on the probe tasks.
 We study these learning curves primarily because they convey precisely the ease or difficulty of navigating the loss landscape. In particular, the learning curve tells us whether optimization is getting trapped in bad minima (in which case the trajectory would hit an early plateau at a large loss value), or whether the network has greater difficulty reducing the loss enough to find a minimum in the first place (corresponding to a flatter slope).
 
 We show in Figure~\ref{fig:learning_curves} the learning curves obtained by an optimization trajectory from parameters $\theta_t$ on the probe task from different timesteps $t$ of training on the RL task. We see that parameters from early training checkpoints quickly attain low losses, but that the slopes of these learning curves become more shallow as training progresses on the main task. Of particular note is the increasing variance of the curves: in the full-batch case, this non-monotonicity is associated with increasing loss landscape sharpness \citep{cohen2021gradient}. In the mini-batch optimization setting, we observed both increasing interference between minibatches as well as non-monotonicity in the loss even on the minibatch on which the gradient was computed. In short, we see that it is increasing difficulty of navigating the loss landscape that drives plasticity loss in this problem.
 
 \section{Solutions}
 \label{sec:experiments}
 Thus far, we have demonstrated that neural networks can lose plasticity even in a task as simple as classifying MNIST digits, assuming that a suitable non-stationarity is introduced into the optimization dynamics. We now turn our attention to means of reducing or reversing this loss of plasticity. Section~\ref{sec:scaling} will evaluate the degree to which scaling alone can eliminate plasticity loss. Section~\ref{sec:interventions} will evaluate the effects of a variety of interventions on plasticity. We test the applicability of these findings to larger scale tasks in Section~\ref{sec:atari}.
 
 \subsection{The role of scaling on plasticity}\label{sec:scaling}
 \begin{figure}
     \centering
     \includegraphics[width=\linewidth]{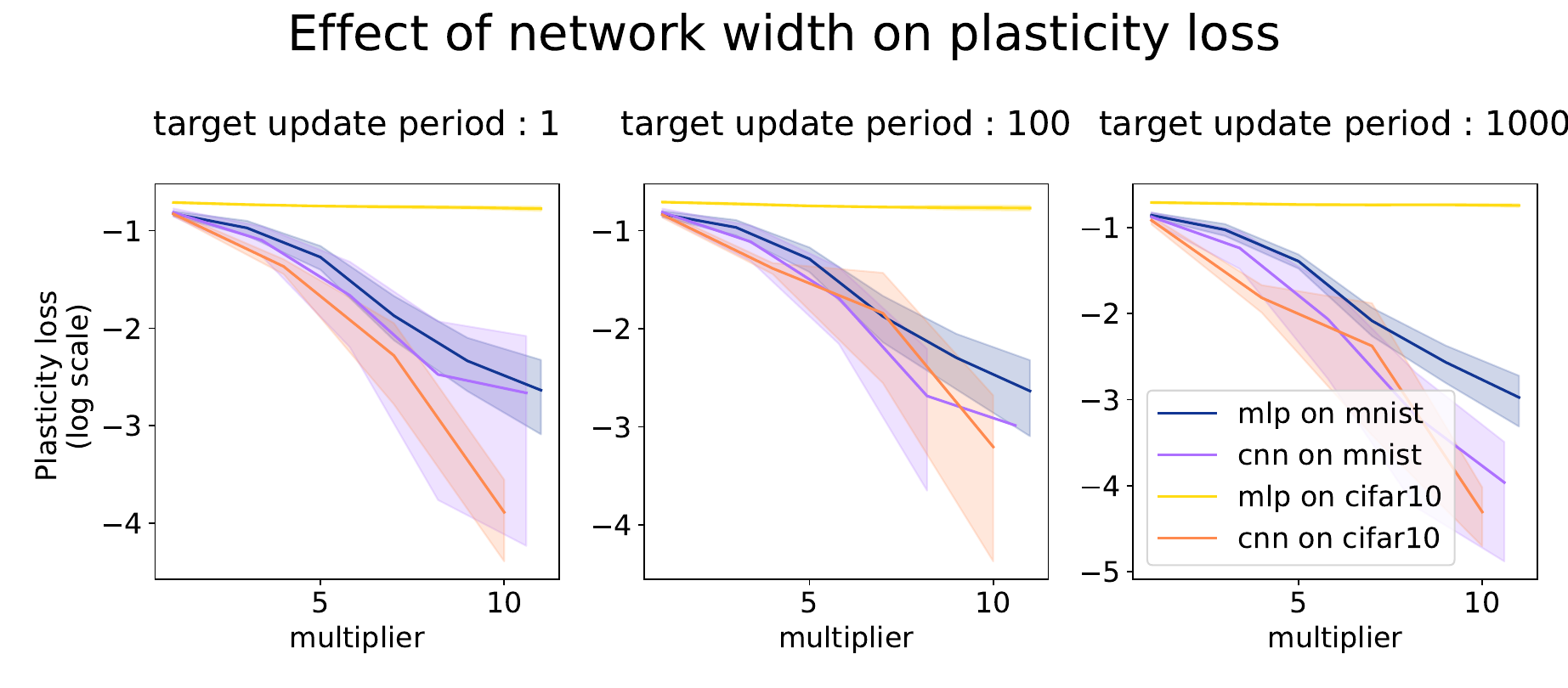}
     \vspace{-1em}
     \caption{We observe a consistent decline in plasticity loss across different target update frequencies as a result of scaling in several architecture-dataset combinations; however, even when scaling the architecture to the point where it no longer fits on a single GPU, we are still unable to completely eliminate plasticity loss on these simple classification-inspired problems.}
     \label{fig:scaling-sweep}
     \vspace{-1em}
 \end{figure}
 Before considering sophisticated methods to address plasticity loss, we must first answer the question of whether this is simply a disease of small networks. In the context of the impressive successes of large models and the resultant scaling law phenomena~\citep{kaplan2020scaling}, it is entirely plausible that plasticity loss, like many other challenges, vanishes in the limit of infinite computation. We find that while plasticity loss is easiest to induce in extreme forms in small networks, scaling a CNN to the limit of a single GPU's memory is insufficient to eliminate plasticity loss even in the simple classification tasks described in the previous section. We visualize the relationship between network width and plasticity loss in Figure~\ref{fig:scaling-sweep}. 
 
 These observations suggest that plasticity loss is unlikely to be the limiting factor for sufficiently large networks on sufficiently simple tasks, particularly when the network is sufficiently overparameterized as to smoothly interpolate its training data \citep{bubeck2023universal}. However, for tasks which do not align with the inductive bias of the network (as in the MLPs trained on CIFAR-10), or for which the network is not sufficiently expressive (as is the case for the small networks of any architecture), we see a reduction in the ability to fit new targets over time. Because we typically cannot guarantee a priori that a learning problem will fall in the first category, we turn our attention to other design choices which might further insure networks against plasticity loss.
 
 \subsection{Interventions in toy problems}\label{sec:interventions}
 
 \begin{figure}
     \centering
     \includegraphics[width=\linewidth]{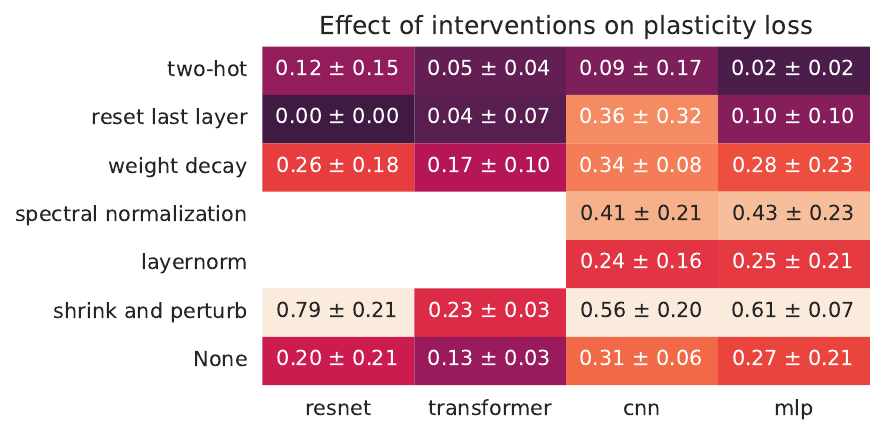}
     \caption{Effect of architectural and optimization interventions on plasticity loss. Colour indicates change in loss on challenge targets between initial and final epoch of training on RL task. Darker shading indicates less plasticity loss.}
     \label{fig:intervention-heatmap}
 \end{figure}
 
 In this section we evaluate the effect of a variety of interventions on plasticity loss. We repeat the protocol used in Section~\ref{subsec:experimental_setting}, training for 100 iterations of 1000 steps. We consider four architectures: a multi-layer perceptron (MLP), a convolutional neural network (CNN) without skip connections, a ResNet-18~\citep{he2016deep}, and a small transformer based on the Vision Transformer (ViT) architecture~\citep{dosovitskiy2020image}.
 
 We consider the following interventions: \textbf{resetting} the last layer of the network at each target network update, a simplified variant of the scheme proposed by \citet{nikishin2022primacy};
 adding \textbf{layer normalization} \citep{ba2016layer} after each convolutional and fully-connected layer of the CNN and the MLP;
 performing \textbf{ Shrink and Perturb} \citep{ash2020warm}: multiplying the network weights by a small scalar and adding a perturbation equal to the weights of a randomly initialized network each time the target network is updated;
 leveraging a \textbf{two-hot} encoding, which presents a distributional formulation of scalar regression wherein the network outputs a categorical probability distribution over fixed support and minimizes a cross-entropy loss with respect to an encoding of a regression target which distributes mass across two adjacent bins of the support;
 \textbf{spectral normalization} of the initial linear layer of the CNN and the MLP~\citep{gogianu2021spectral}; and
 \textbf{weight decay,} setting the $\ell_2$ penalty coefficient to $10^{-5}$.
 
 These methods were chosen to be representative samples of a number of approaches to mitigating plasticity loss: resetting the last layer temporarily removes a source of poor conditioning from the optimization process while likely not significantly influencing training dynamics elsewhere \citep{zhang2019all}; layer normalization and residual connections tend to make networks more robust to optimizer choices; weight decay and spectral normalization both regularize the parameters of the network in different ways; shrink and perturb applies a perturbation to the current parameters without significantly changing the decision boundary (though we note that for regression tasks this will still influence the scale of the network outputs, and so may not be desirable).
 
 We visualize our key takeaways in Figure~\ref{fig:intervention-heatmap}, which compares plasticity loss after 100 iterations of training on each of the architecture-intervention combinations. 
 Overall, selecting a network parameterization which smooths out the loss landscape is the most effective means of preserving plasticity of all approaches we have considered in this setting, and even has a greater effect on plasticity than resetting the final layer of the network in some instances. We visualize some learning curves of networks with and without layer normalization in Figure~\ref{fig:cnn-easy} in the supplementary material.
 
 We note that while the two-hot encoding does demonstrate significant reductions in plasticity loss, it does so at the cost of stability of the learned policy in several instances we considered. Additionally, this intervention required significantly different optimizer hyperparameters from the regression parameterization, suggesting that while it can be a powerful tool to stabilize optimization, it might not be suitable as a plug-in solution to mitigate plasticity loss in an existing protocol.
 
 \subsection{Application to larger benchmarks}
 \label{sec:atari}
 
 \begin{figure}[t]
     \centering
     \includegraphics[width=0.9\linewidth]{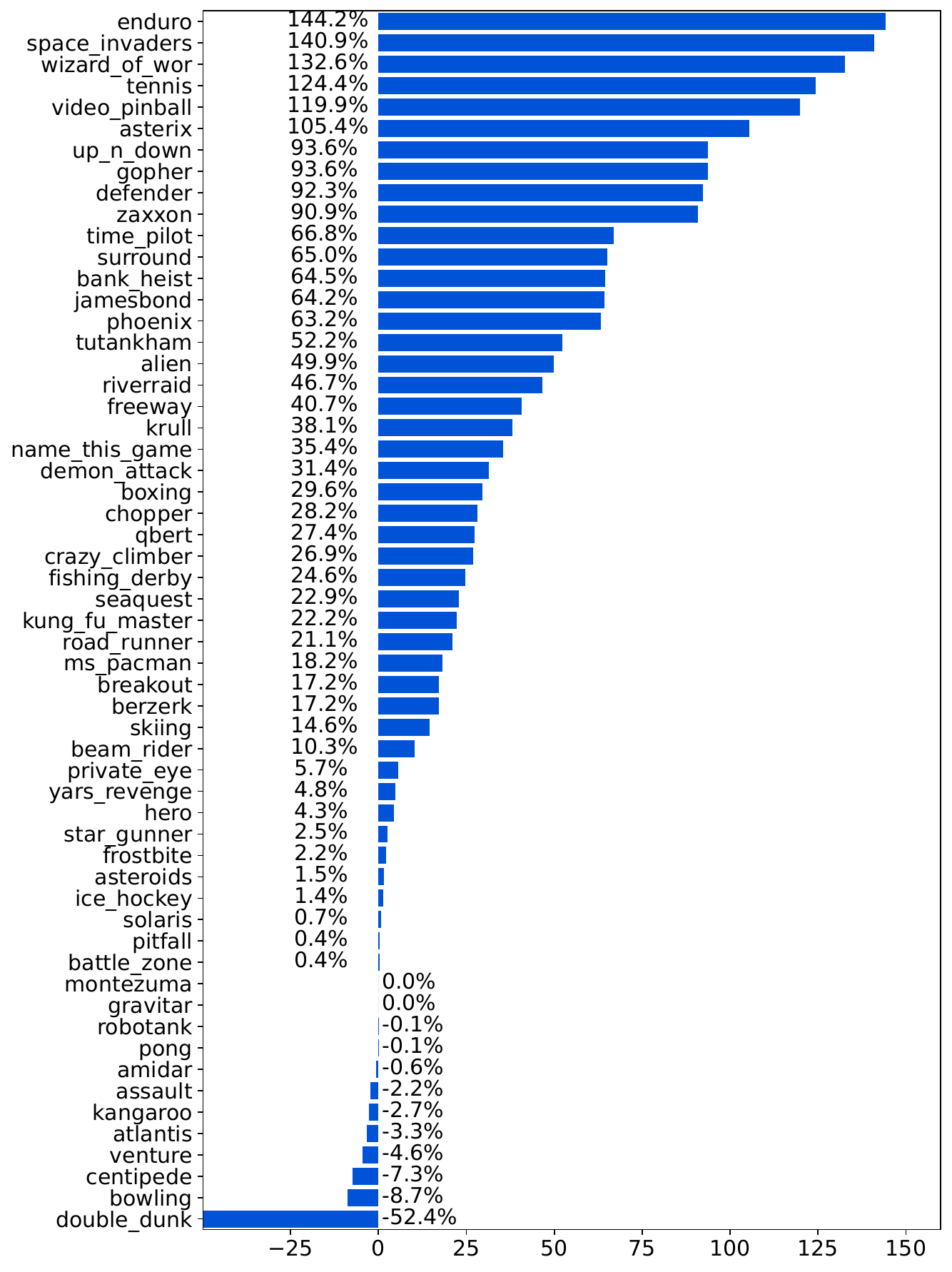}
     \includegraphics[width=0.47\linewidth]{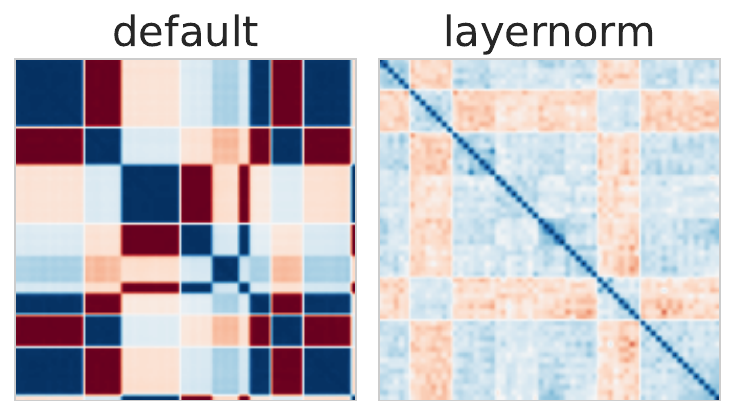}
     \vline
     \includegraphics[width=0.47\linewidth]{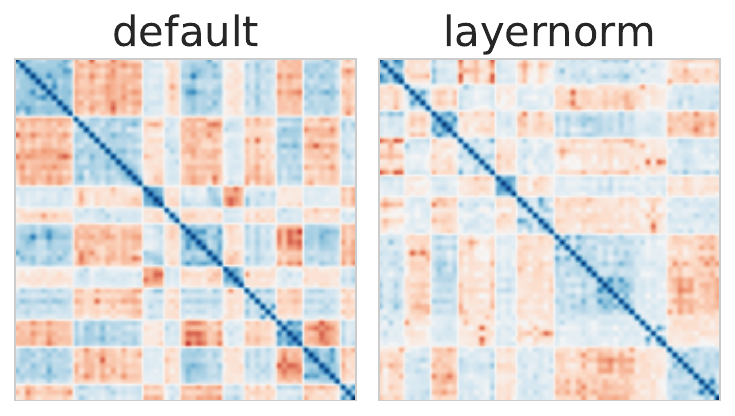}
     \caption{Layer normalization improves performance and changes the gradient covariance structure in DDQN agents. Top: Human-normalized improvement score~\citep{wang2016dueling} of adding layer normalization over the default double DQN agent. Bottom: Gradient covariance matrices for Freeway (left) and Kangroo (right). In environments where layer normalization significantly improves performance, it also induces weaker gradient correlation.}
     \vspace{-1em}
     \label{fig:atari}
 \end{figure}
 
 We now evaluate whether the benefits of layer normalization on plasticity in toy classification tasks translate to larger-scale benchmarks.
 We use the standard implementation of double DQN~\citep{van2016deep} provided by~\citet{dqnzoo2020github}, and evaluate three seeds on each of the 57 games in the Arcade Learning Environment benchmark~\citep{bellemare2013arcade}. We use the RMSProp optimizer, $\epsilon$-greedy exploration, and frame stacking~\citep{mnih2015human}. Full implementation details can be found in Appendix~\ref{sec:atari-details}. The only difference between the baseline implementation and our modification is the incorporation of layer normalization after each hidden layer in the network.
 
 We see in Figure~\ref{fig:atari} that the introduction of layer normalization robustly improves performance across the benchmark, without any additional hyper parameter tuning. While this improvement cannot be definitively attributed to a reduction in plasticity loss from the evidence provided, it points towards the regularization of the optimization landscape as a fruitful direction towards more robust RL agents. 
 We further observe that many of the environments where layer normalization offers a significant boost to performance are those where the gradient covariance structure of the default architecture is degenerate or where the Hessian is ill-conditioned, and the LN networks which obtain performance improvements tend to have correspondingly better behaved gradient covariance. We provide a hint into this phenomenon in Figure~\ref{fig:atari}, and defer the complete evaluation over all 57 games to Appendix~\ref{sec:qualitative-dqn}.
 
 \section{Related Work}
 
 \textbf{Trainability:} the problem of finding trainable neural network initializations is well-studied \citep{glorot2010understanding, he2015delving, sutskever2013importance}. Without careful initialization and architecture design, vanishing and exploding gradients may arise \citep{yang2017mean}. ResNets \citep{he2016deep} in particular are known to resolve many of these pathologies by biasing each layer's mapping towards the identity function, leading to better-behaved gradients \citep{balduzzi2017shattered}. Mean-field analysis \citep{yang2017mean, schoenholz2017deep, yang2018a}, information propagation \citep{poole2016exponential}, and deep kernel shaping \citep{zhang2021deep, martens2021rapid} have all been applied to study trainability in neural networks, and to characterize the role of residual connections. A wealth of prior work additionally studies the role of loss landscape smoothness in generalization and performance \citep{li2018visualizing, shibani2018how, ghorbani2019investigation, park2022how}. Other works highlight the chaotic behaviour of early training periods \citep{Jastrzebski2020The}, in particular the `edge of stability' phenomenon \citep{cohen2021gradient} and the `catapault mechanism' \citep{lewkowycz2020large},  and relate closely to the observations grounding `linear mode connectivity' \citep{frankle2020linear} to explain generalization and trainability in deep neural networks; however, these approaches all focus on supervised learning with a stationary objective.
 
 \textbf{Continual learning} encompasses a broad range of problem settings \citep{berariu2021study, hadsell2020embracing, rolnick2019experience} encompassing both input distribution shift and covariate shift. The connection between plasticity and non-stationarity was first observed in the case of the former \citep{ash2020warm}; however this paper has focused on the latter as a driver of performance plateaus in RL.
 Most related to our study is the identification of the \textit{loss} of plasticity as a potentially limiting factor in deep reinforcement learning \citep{lyle2021understanding, dohare2021continual}. This study can be motivated by the rich literature studying the effect of resetting and distillation on performance \citep{fedus2020catastrophic, nikishin2022primacy, igl2021transient, schmitt2018kickstarting}. 
 

 \section{Conclusions}
 The findings of this paper highlight a stark contrast between recent observations on pretraining in large models, which find that suitable learning objectives can accelerate adaptation and improve generalization on later tasks, and the loss of plasticity in non-stationary prediction problems, where \textit{un}suitable objectives can hurt a network's ability to adapt to new learning signals. However, as reinforcement learning algorithms scale up to more complex tasks, the divide between these regimes shrinks. While it is possible that in many settings, plasticity loss is not a limiting factor in network performance and so need not be a concern for many of the relatively small environments used to benchmark algorithms today, we conjecture that as the complexity of the tasks to which we apply RL grows, so will the importance of preserving plasticity.
 
 The findings of this paper point towards stabilizing the loss landscape as a crucial step towards promoting plasticity. This approach is likely to have many ancillary benefits, presenting an exciting direction for future investigation. A smoother loss landscape is both easier to optimize and tends to exhibit better generalization, and it is an exciting direction for future work to better disentangle the complementary roles of memorization and generalization in plasticity.
 
 \section*{Acknowledgements}
 We thank Mark Rowland, Tom Schaul, Georg Ostrovski, Hado van Hasselt, Diana Borsa, and Samuel L Smith for valuable feedback and discussions during the development of this work.
 \newpage
 \bibliography{example_paper}
 \bibliographystyle{icml2023}

 \newpage
 \appendix
 \onecolumn
 
\section{Experiment details}
\subsection{Case studies}
\label{sec:case-study-details}
\textbf{Optimizer instability:} we consider a memorization problem on the MNIST dataset \citep{deng2012mnist}, where the network is trained to classify its inputs according to randomly permuted labels of a subset of 5000 MNIST images. We use a fully-connected multi-layer perceptron (MLP) with two hidden layers of width 1024. At the end of each training iteration we log the accuracy on a sample of 4096 states, and re-randomize the labels, keeping the input set fixed. We train the network with an adam optimizer with learning rate equal to 0.001, first-order moment decay $b_1=0.9$, second-order moment decay $b_2=0.999$, $\varepsilon = 10^{-9}$, and $\bar{\varepsilon}=0$. With the tuned optimizer, we se $b_2=0.9$ and $\bar{\varepsilon}=10^{-3}.$

\textbf{Brownian motion:} we train the network via a Q-learning loss on batches of transitions generated by the easy classification MDP with the MNIST observation space. We use a stochastic gradient descent optimizer with learning rate 0.001. We use a batch size of 512. 

We evaluate the gradient covariance matrix $C_k$ by sampling a batch of transitions and computing the gradient of each transition individually. We then take the normalized dot product matrix and permute the rows and columns according to a k-means clustering, where we set $k=10$ to match the number of latent states in the environment. We update the target network in the Q-learning objective once every 5000 steps.

To compute the Hessian eigenvalue density, we follow the implementation of \citet{ghorbani2019investigation} in order to obtain a Gaussian approximation to the eigenvalue distribution. We sample a single large batch of transitions, and use the Lanczos algorithm to obtain a set of centroids which are then convolved with a Gaussian distribution to obtain the final density.
 
\subsection{Toy RL environments}
\label{sec:toy-details}
We use the same agent structure for both the Atari and classification MDP environments. The agent collects data by interacting with the environment following an $\epsilon$-greedy policy and stores states in a replay buffer. We then interleave interaction with the environment and optimization on sampled batches from the replay buffer. In the classification MDP, we train the network for 10,000 steps before updating the target network. We tried a variety of target network update frequencies and found that shorter update periods resulted in poor action-value estimation in the hard environment. We probe the ability of the network to fit a new set of regression targets once every 5000 optimizer steps: we draw 10 randomly sampled target functions generated by the procedure described in Section~\ref{sec:plasticity-definition}, and for each run the network's optimizer from the current parameters to minimize the loss with respect to these new targets for 2000 steps.  

The architectures we consider in our plasticity evaluations are as follows:
\begin{itemize}
    \item MLP: we use two hidden layers of varying width. For all evaluations other than the width sweep used to generate Figure~\ref{fig:scaling-sweep}, we use a width of 512. For the width sweep, we set a base width of 16 and then multiply by factors of 1, 2, 4, 8, 12, and 16. 
    \item CNN: we use two convolutional layers followed by two fully-connected layers. The first convolutional layer uses 5x5 kernels, while the second uses 3x3; both have 64 channels. The fully-connected layers have widths of 256.
    \item ResNet: we use a standard resnet18 architecture \citep{he2016deep}.
    \item Vision Transformer: we use our own implementation based on \citep{dosovitskiy2020image}. We use a patch size of 3 to construct the convolutional embeddings, model dimension of 256 and a feedforward width of 1024. We use a single transformer block, and a dropout rate of 0.1. All components of the model are trained from scratch on the task.
\end{itemize}

\subsection{Double DQN}
\label{sec:atari-details}
We follow the standard training protocol on Atari, training for 200 million frames and performing optimizer updates once every 4 environment steps \citep{dqnzoo2020github}. We add layer normalization after each hidden layer of the network. We use a replay buffer of size 100,000, and follow an $\epsilon$-greedy policy during training with $\epsilon=0.1$.

\section{Additional analysis}
\subsection{Detailed intervention analysis}
\begin{figure}
    \centering
    \includegraphics[width=0.33\linewidth]{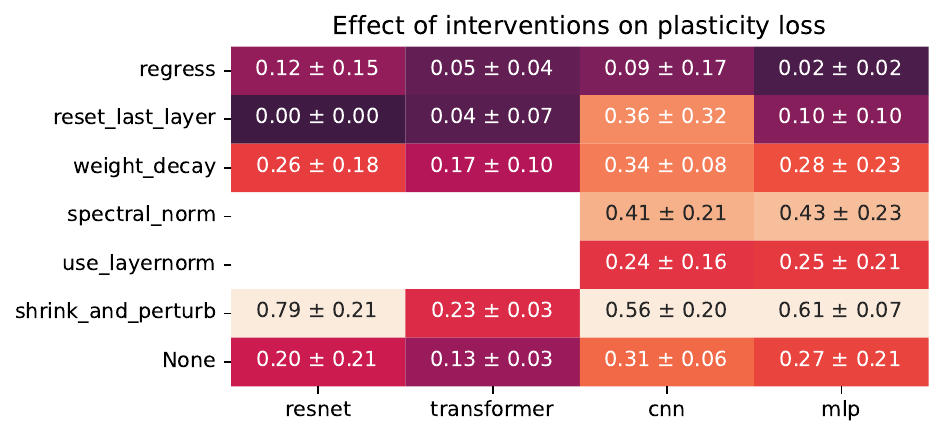}
    \includegraphics[width=0.32\linewidth]{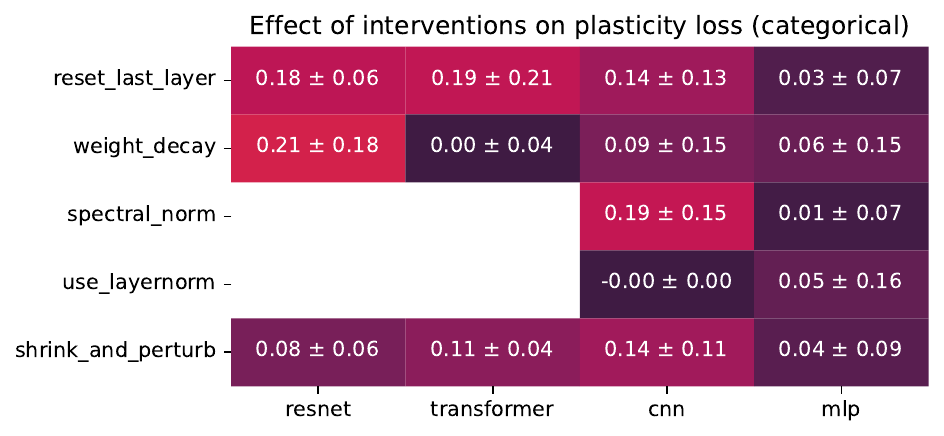}
    \includegraphics[width=0.32\linewidth]{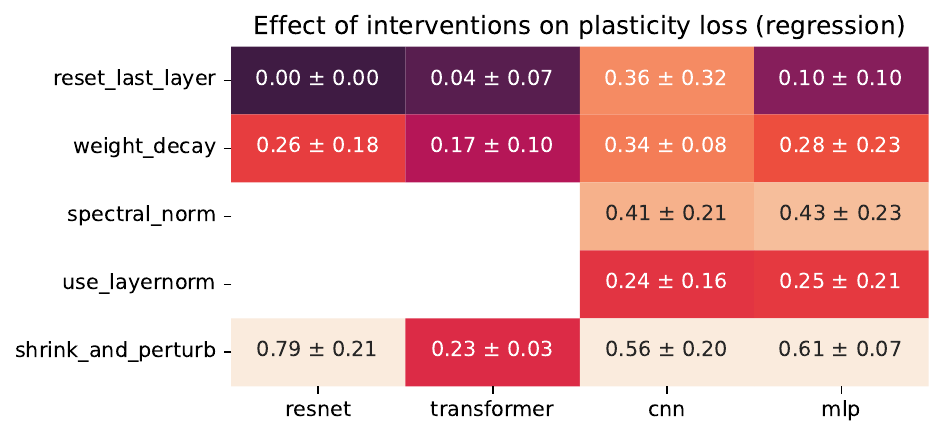}
    \caption{We repeat the analysis of Figure~\ref{fig:intervention-heatmap}, but also include the effect of interventions on regression and categorical output encodings. We observe a significant benefit in the transformer, CNN, and MLP architectures from using the categorical encoding. This figure shows results for the CIFAR-10 dataset.}
    \label{fig:detailed-interventions-cifar}
\end{figure}

\begin{figure}
    \centering
    \includegraphics[width=0.33\linewidth]{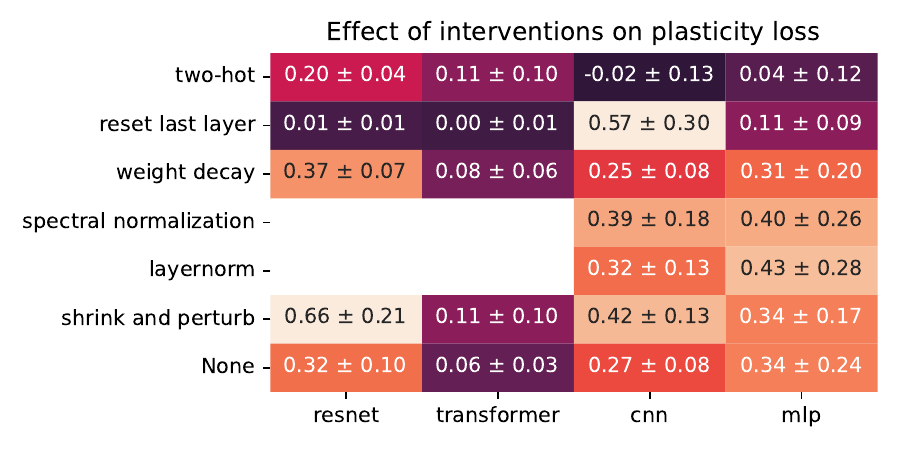}
    \includegraphics[width=0.32\linewidth]{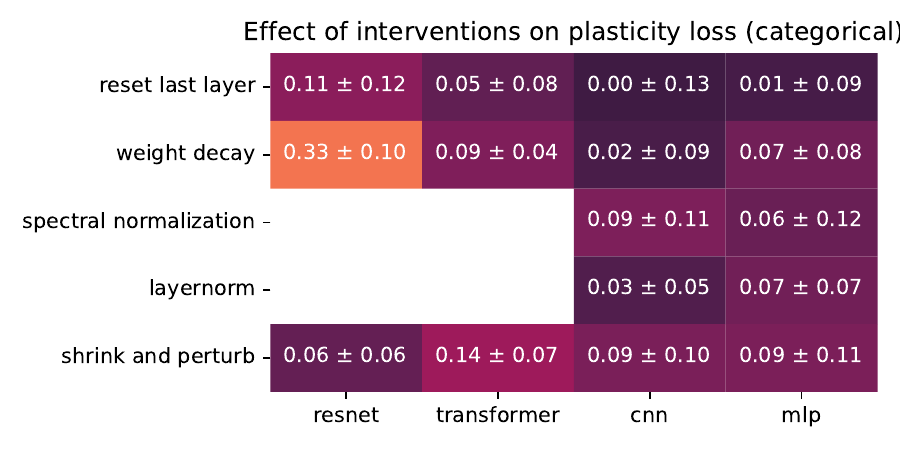}
    \includegraphics[width=0.32\linewidth]{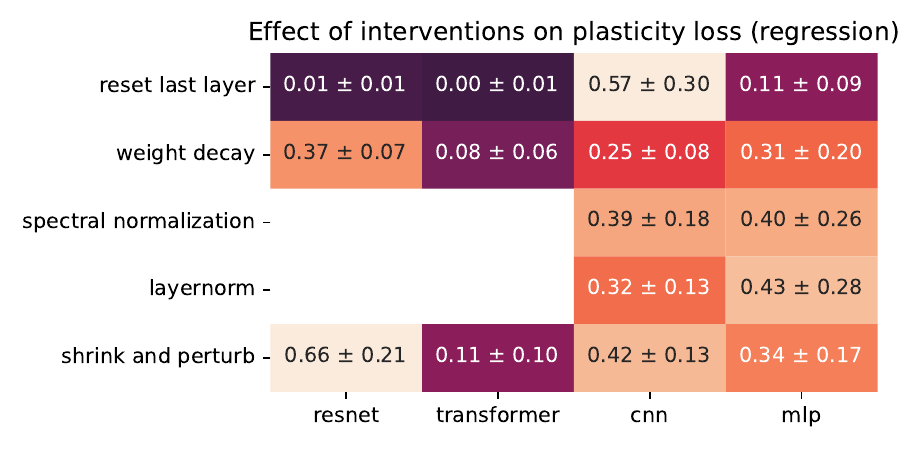}
    \caption{We repeat the analysis of Figure~\ref{fig:intervention-heatmap}, but also include the effect of interventions on regression and categorical output encodings. We observe a significant benefit in the transformer, CNN, and MLP architectures from using the categorical encoding.}
    \label{fig:detailed-interventions-mnist}
\end{figure}

We provide a more detailed analysis of the effects of a variety of interventions on plasticity loss in different neural network architectures. Figures~\ref{fig:detailed-interventions-cifar} and~\ref{fig:detailed-interventions-mnist} show the \textit{change} in the average final probe task loss after training on the toy RL environments for 1 million optimizer steps. We see that resetting the last layer, incorporating a two-hot output representation, and performing layer normalization have beneficial effects on plasticity, whereas shrink and perturb, weight decay, and resetting only the optimizer state do not improve plasticity.

\begin{figure}
    \centering
    \includegraphics[width=0.33\linewidth]{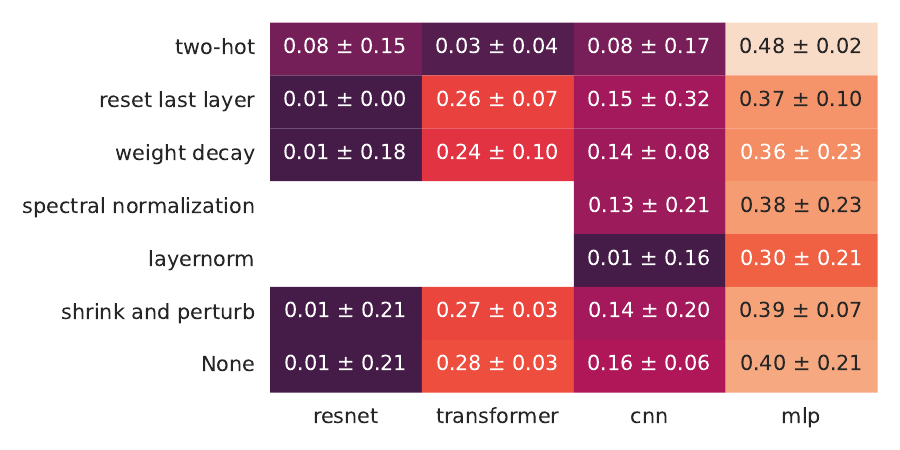}
    \includegraphics[width=0.32\linewidth]{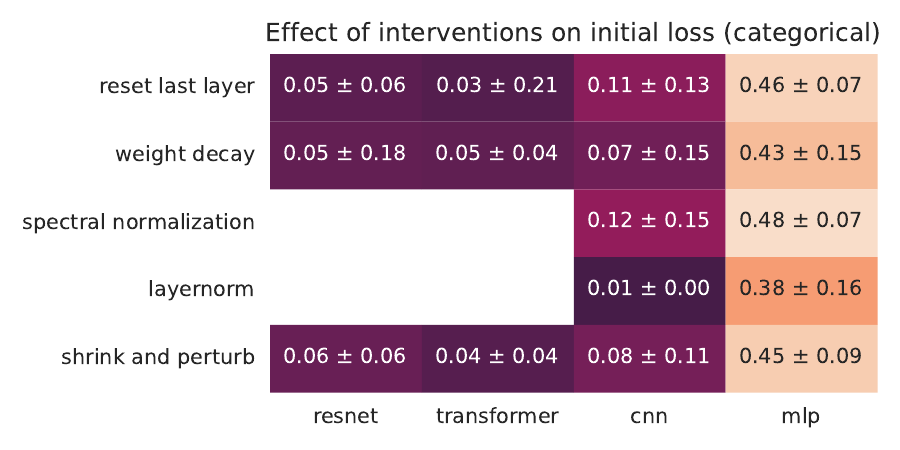}
    \includegraphics[width=0.32\linewidth]{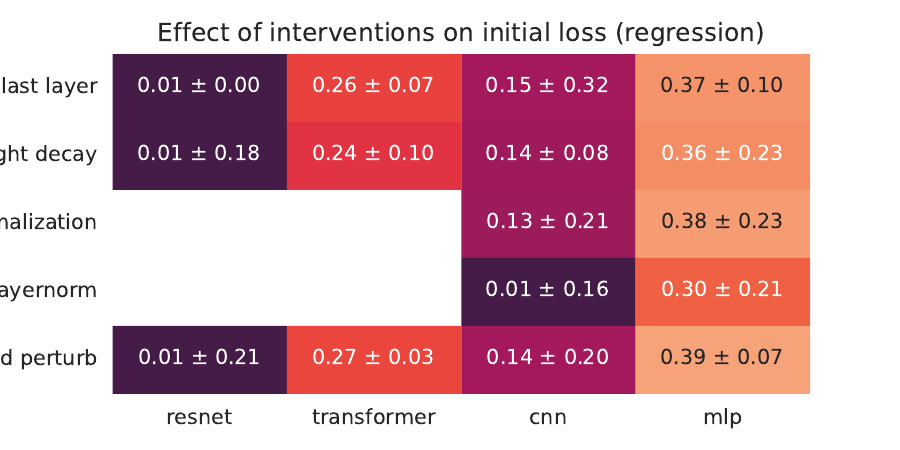}
    \includegraphics[width=0.33\linewidth]{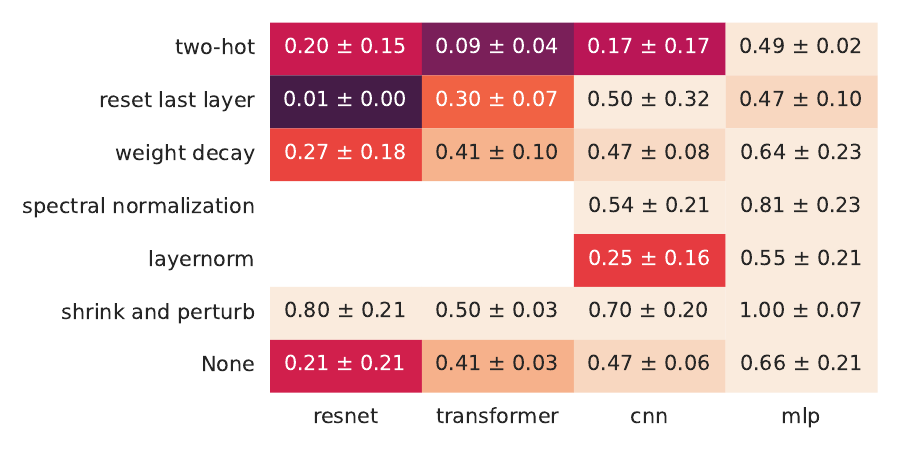}
    \includegraphics[width=0.32\linewidth]{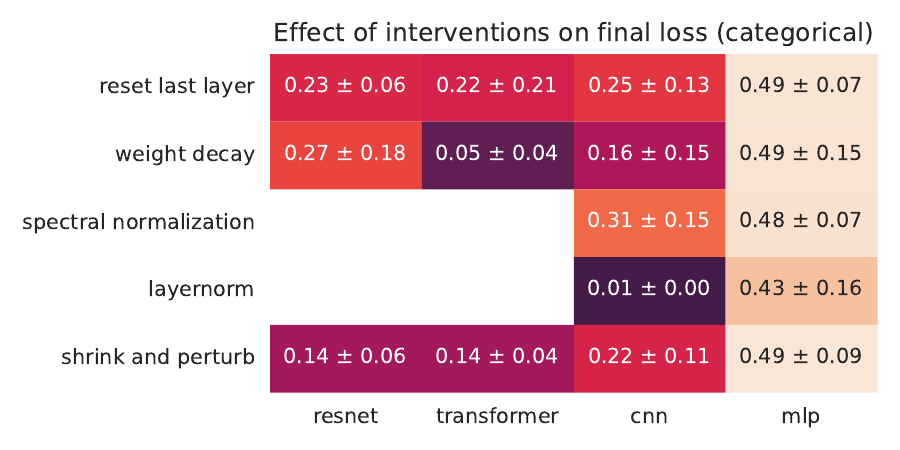}
    \includegraphics[width=0.32\linewidth]{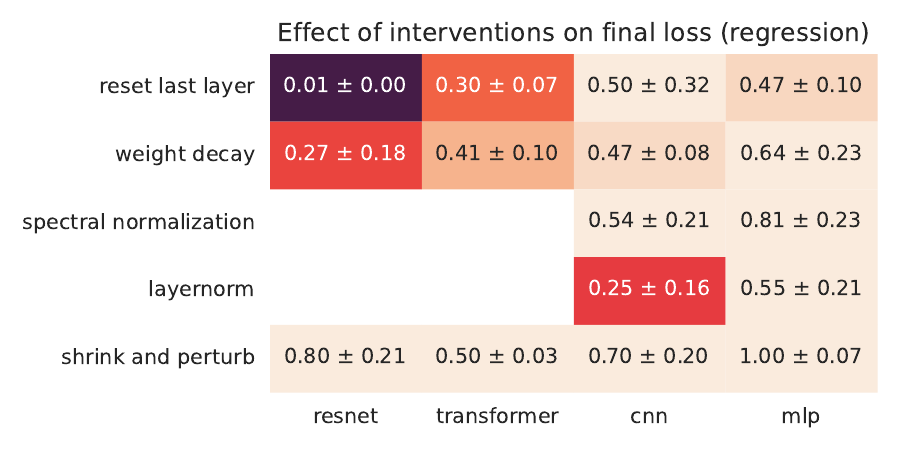}
    \caption{Initial and final loss evaluation on CIFAR-10 input distribution.}
    \label{fig:detailed-interventions-initial-final-cifar}
\end{figure}

\begin{figure}
    \centering
    \includegraphics[width=0.33\linewidth]{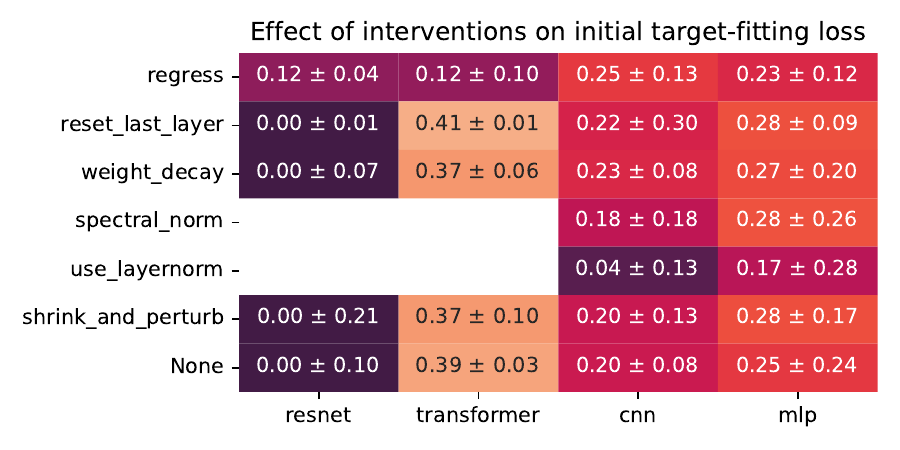}
    \includegraphics[width=0.32\linewidth]{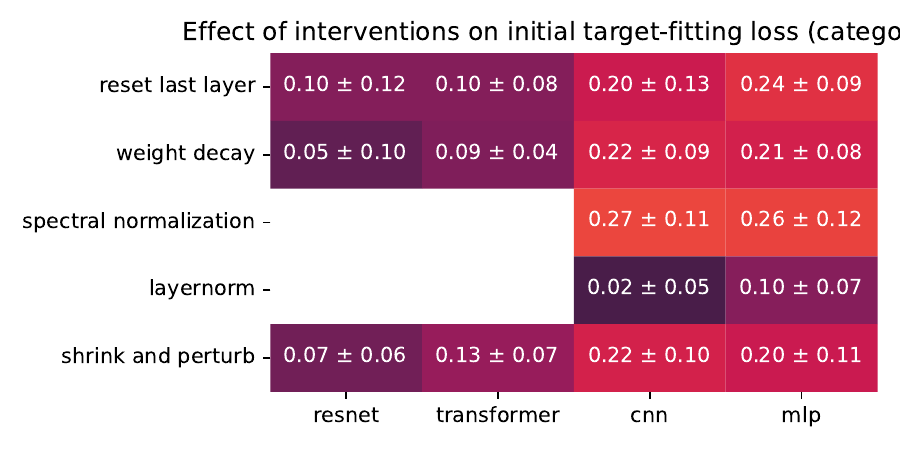}
    \includegraphics[width=0.32\linewidth]{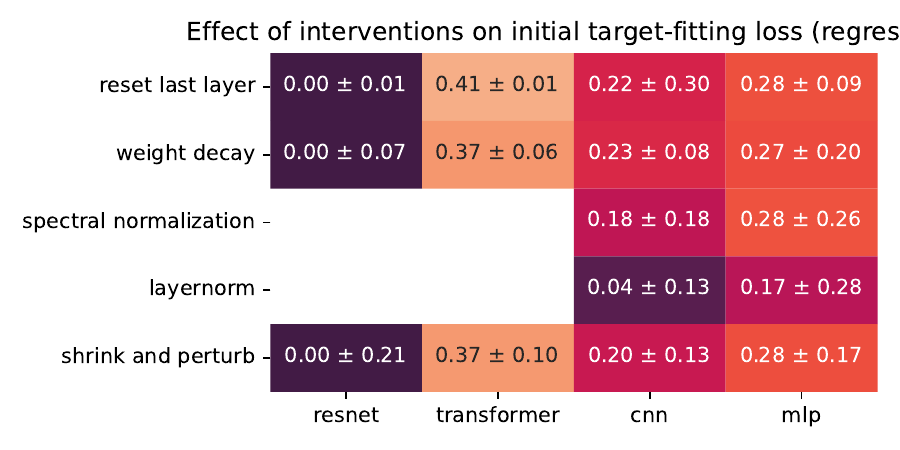}
    \includegraphics[width=0.33\linewidth]{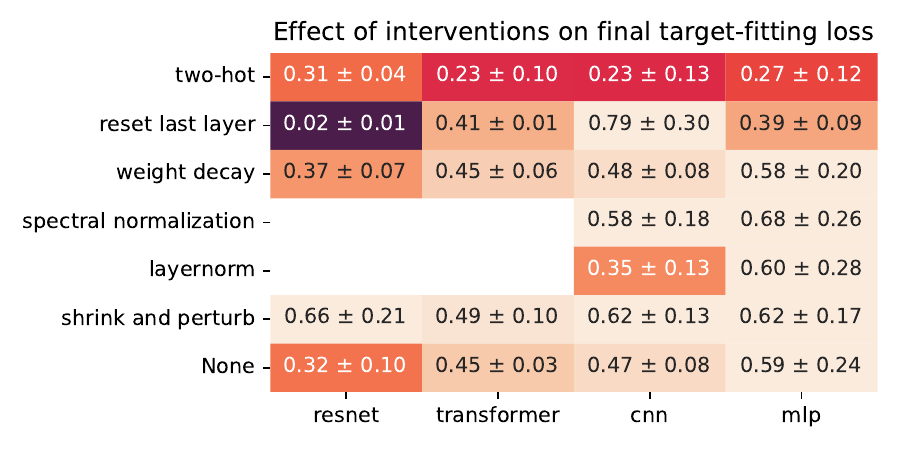}
    \includegraphics[width=0.32\linewidth]{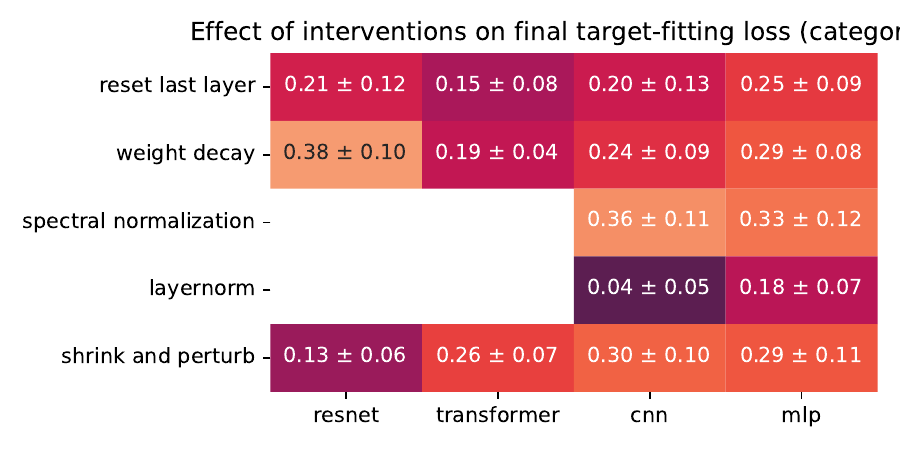}
    \includegraphics[width=0.32\linewidth]{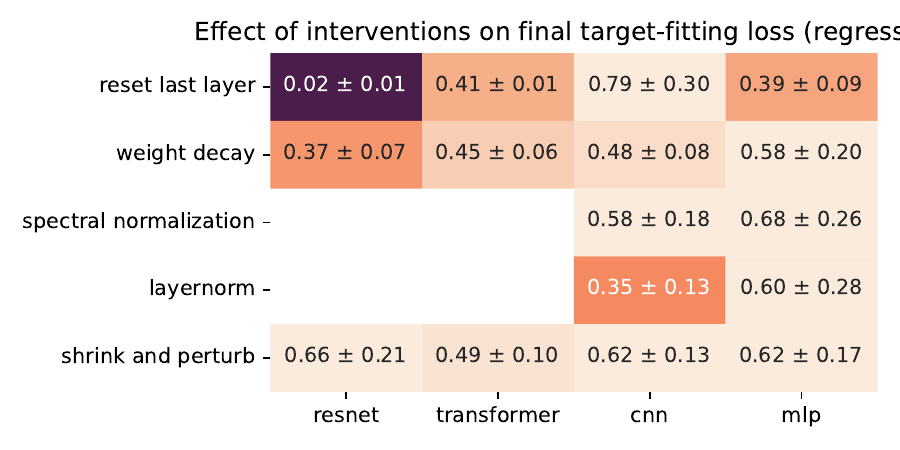}
    \caption{Initial and final loss evaluation on MNIST input distribution.}
    \label{fig:detailed-interventions-initial-final-mnist}
\end{figure}

In Figures~\ref{fig:detailed-interventions-initial-final-cifar} and \ref{fig:detailed-interventions-initial-final-mnist} we show the effect of each intervention scheme on the initial and final losses. Some interventions improve trainability even from initialization, for example using a categorical output for the transformer, in addition to reducing the final probe task loss. The categorical representation appears to combine nicely with other interventions such as layer normalization, resetting, and shrink and perturb. The benefits of the categorical parameterization on shrink and perturb are expected, as regression targets are not output-scale-invariant in the same way that softmax logits are. 

\begin{figure}
    \centering
    \includegraphics[width=0.485\linewidth]{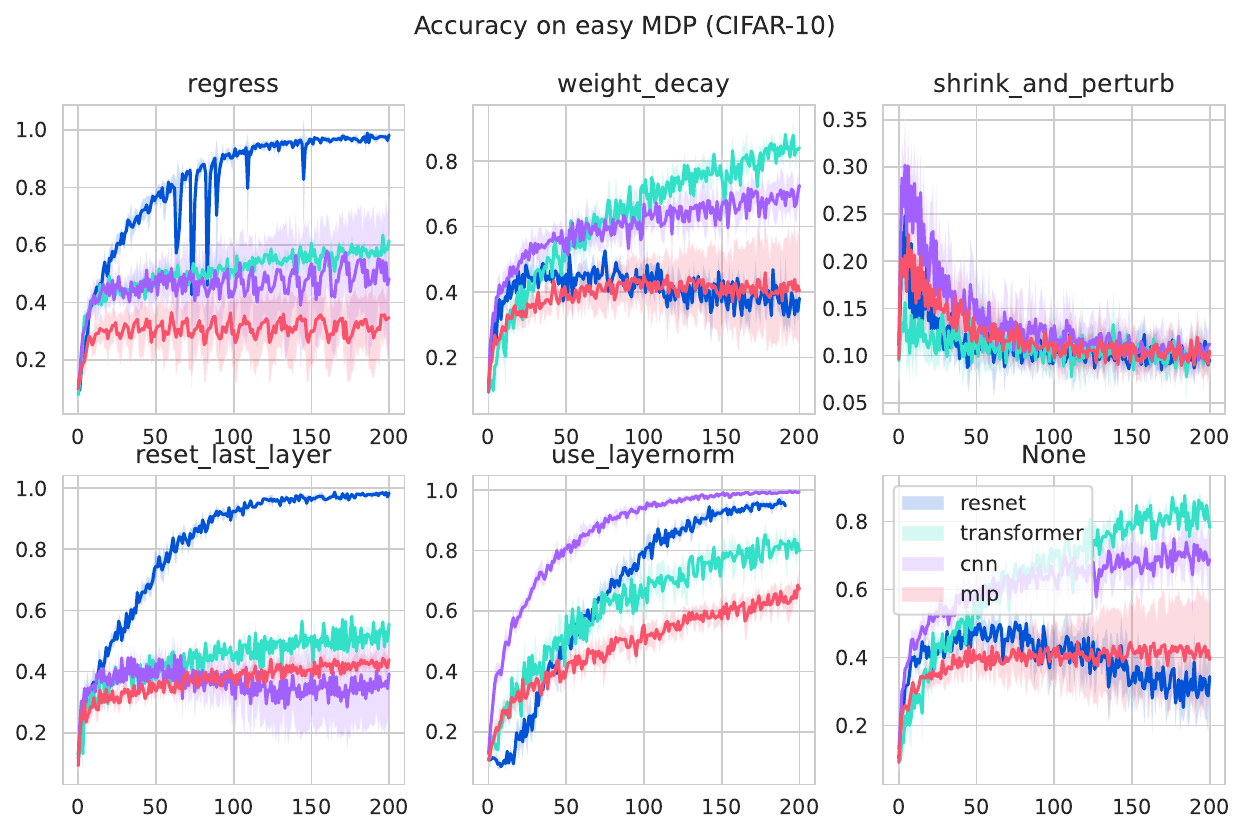}
    \includegraphics[width=0.485\linewidth]{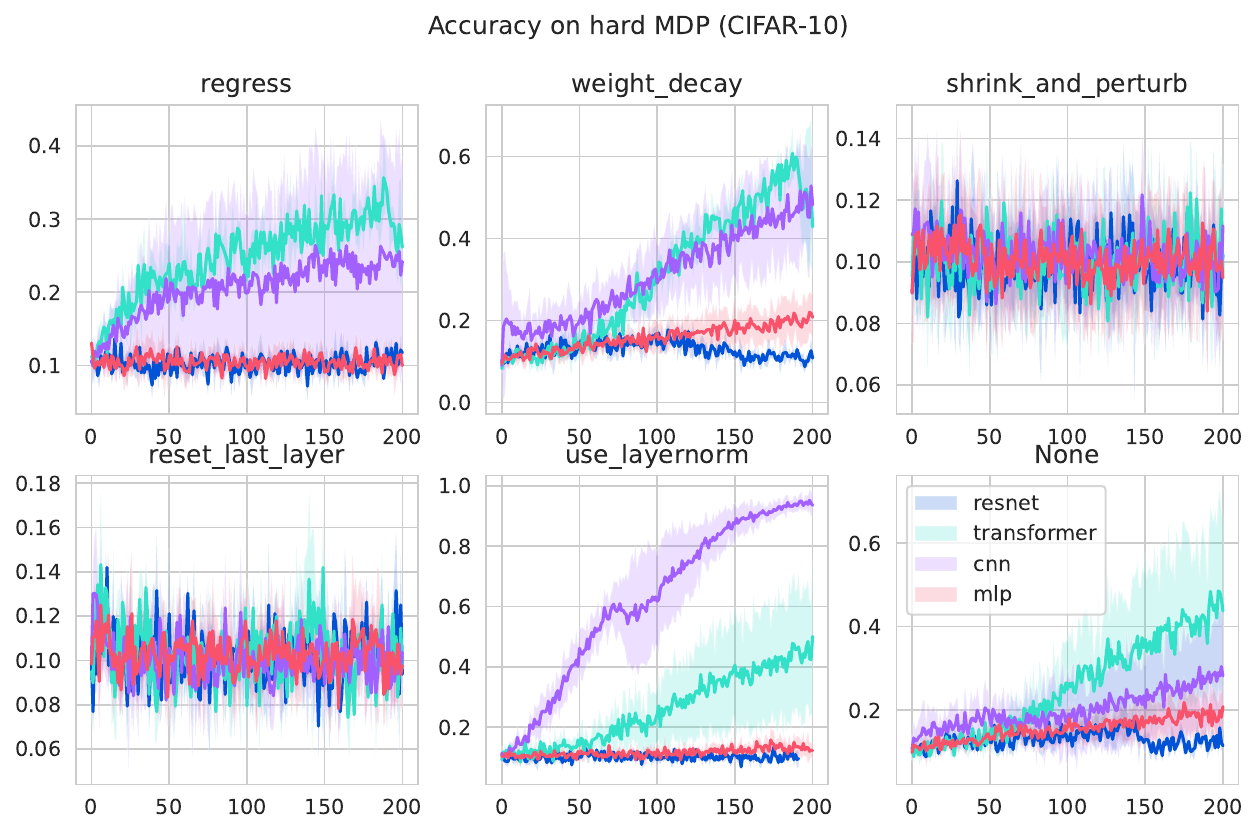}
    
    \caption{We visualize agent performance on the `easy' and `hard' tasks on MDPs using the CIFAR-10 observation space.}
    \label{fig:accuracies-interventions-cifar}
\end{figure}

\begin{figure}
    \centering
    \includegraphics[width=0.485\linewidth]{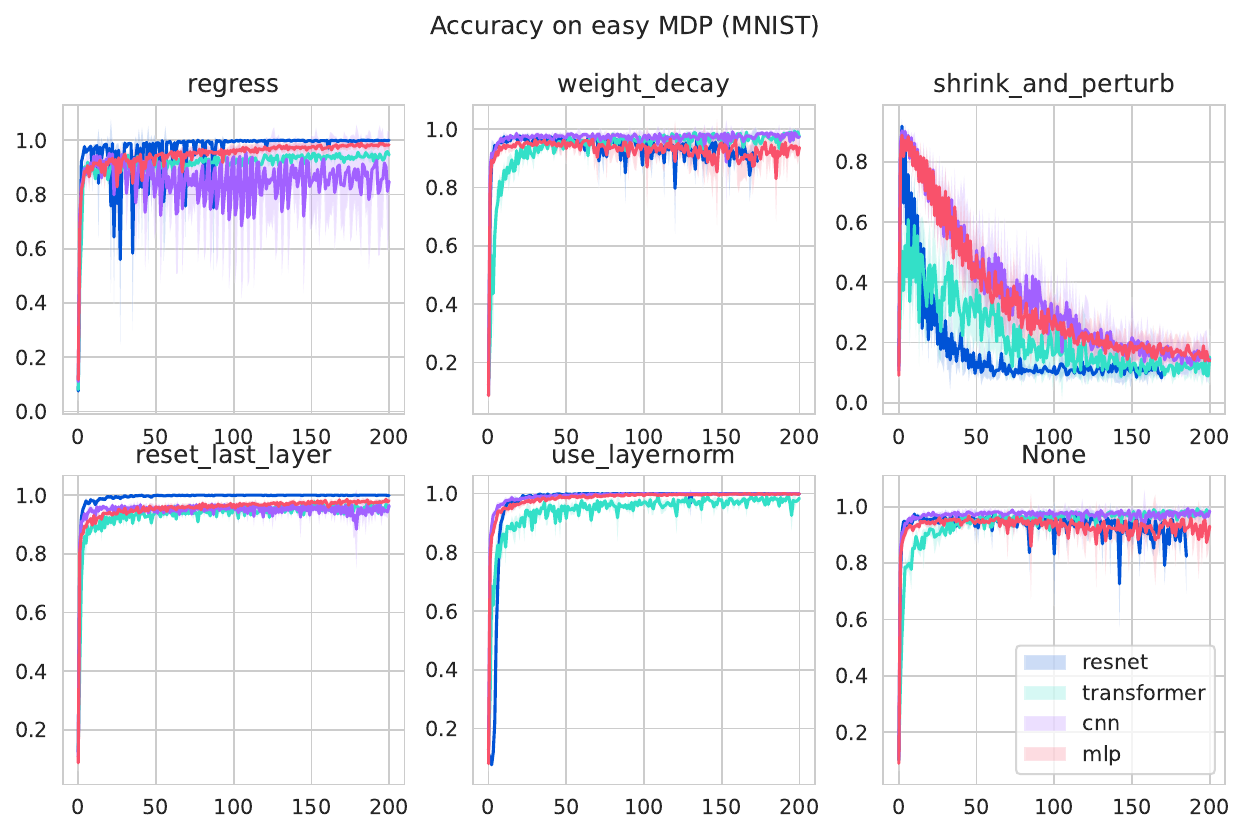}
    \includegraphics[width=0.485\linewidth]{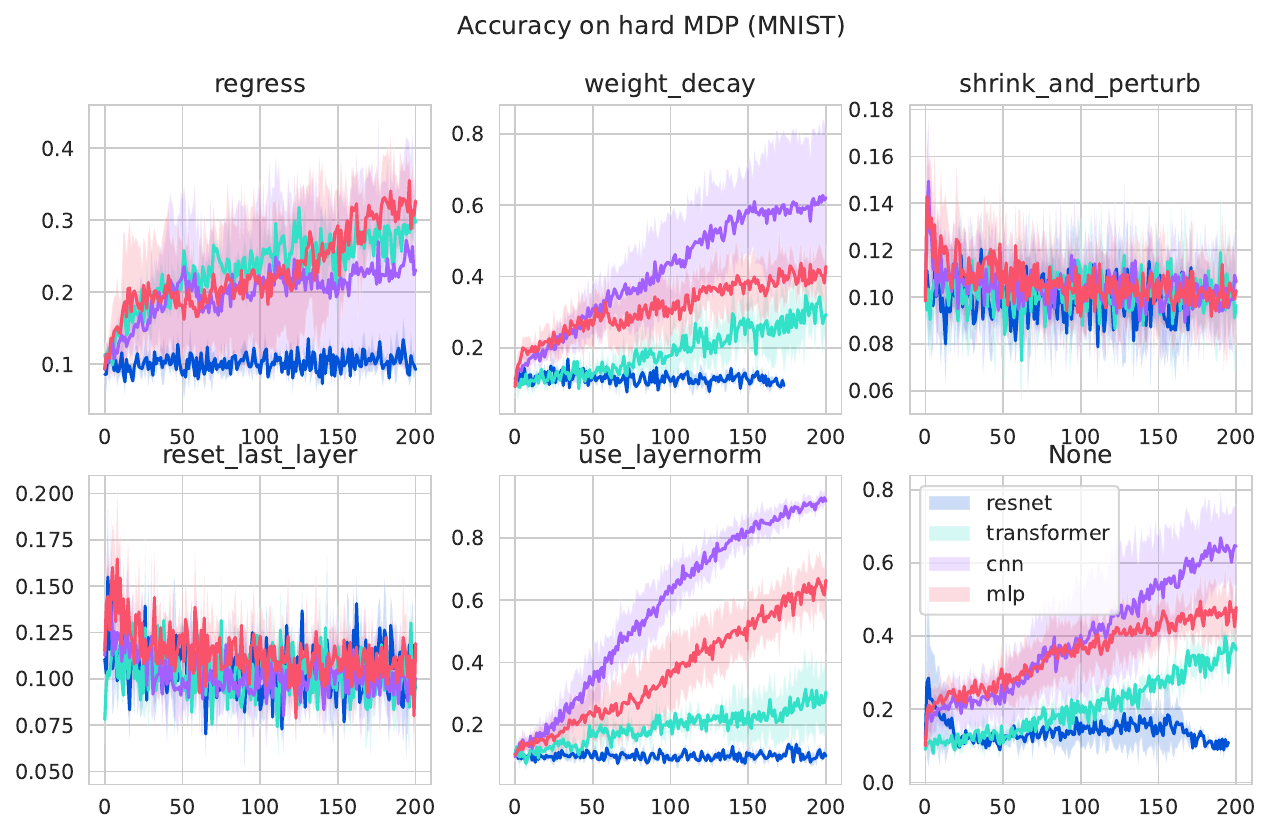}
    
    \caption{We visualize agent performance on the `easy' and `hard' tasks on MDPs using the MNIST observation space.}
    \label{fig:accuracies-interventions-mnist}
\end{figure}

We additionally validate whether these interventions have the potential to interfere with learning on the primary task of interest in Figures~\ref{fig:accuracies-interventions-cifar} and \ref{fig:accuracies-interventions-mnist}. We observe that the methods which perturb the network weights often interfere with learning, particularly in the more challenging `hard' reward structure that requires training for several tens of thousands of steps to make performance improvements. We note that we did not perform extensive hyperparameter tuning for each intervention, so it is possible that are more carefully tuned optimizer would produce better performance. Instead, these results should be taken holistically as an indicator of the robustness of an intervention to using a reasonable optimizer as was used to train the original network to which it is being applied.

\begin{figure}
    \centering
    \includegraphics[width=0.3\linewidth]{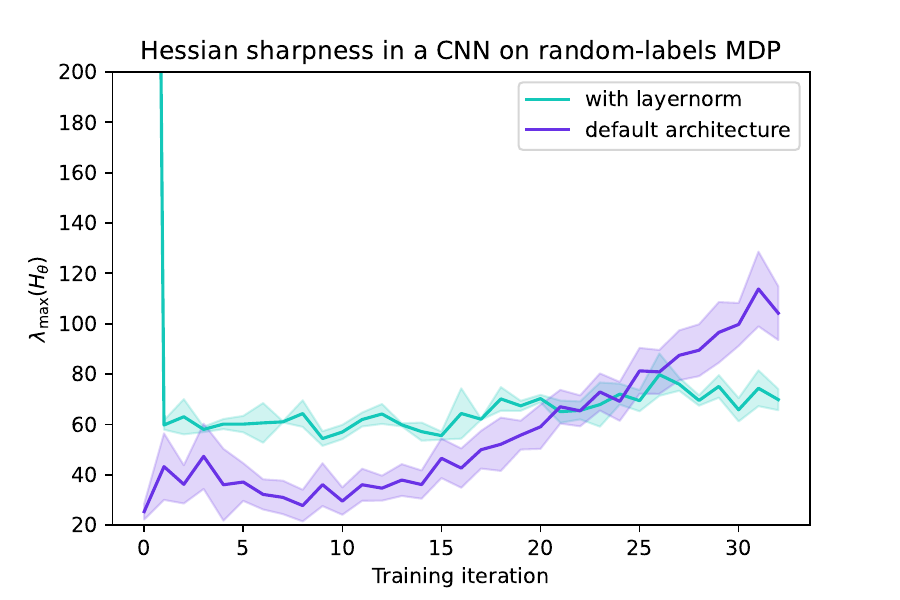}
    \includegraphics[width=0.3\linewidth]{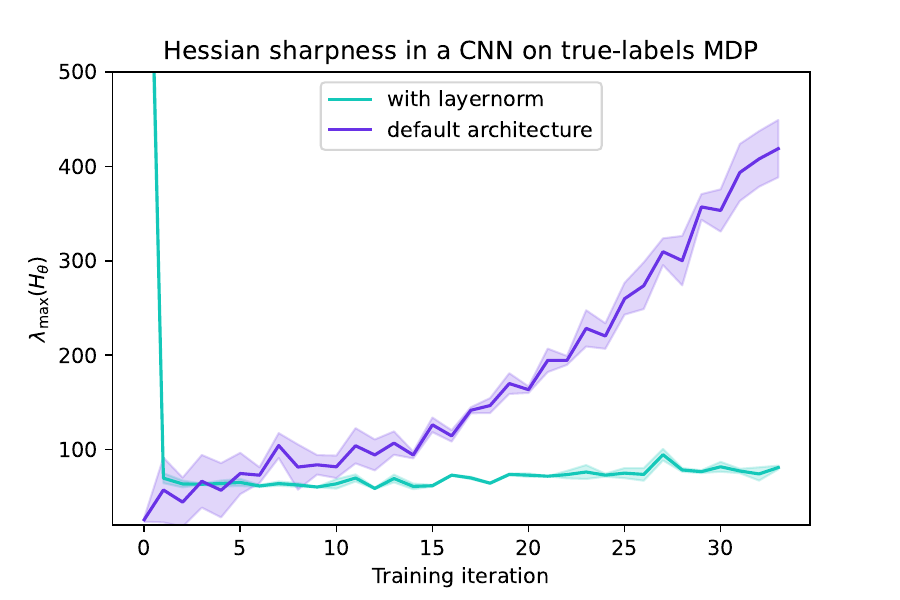}
    \caption{Layer normalization stabilizes the spectral norm of the Hessian in convolutional networks trained with regression on the classification MDP tasks. The network with larger output magnitudes (right) exhibits a greater smoothing effect from layer normalization.}
    \label{fig:layernorm-hessian}
\end{figure}

Finally, we observe that in the convolutional architecture layer normalization often provides a significant stabilization effect on the sharpness of the loss landscape. We observe in the convolutional network trained via a regression loss that the added layer normalization prevents an increase in sharpness that occurs in the default architecture, as visualized in Figure~\ref{fig:layernorm-hessian}.
\subsection{Learning curves for classification MDPs}
\subsubsection{Training accuracy}
\begin{figure}
    \centering
    \includegraphics[width=0.485\linewidth]{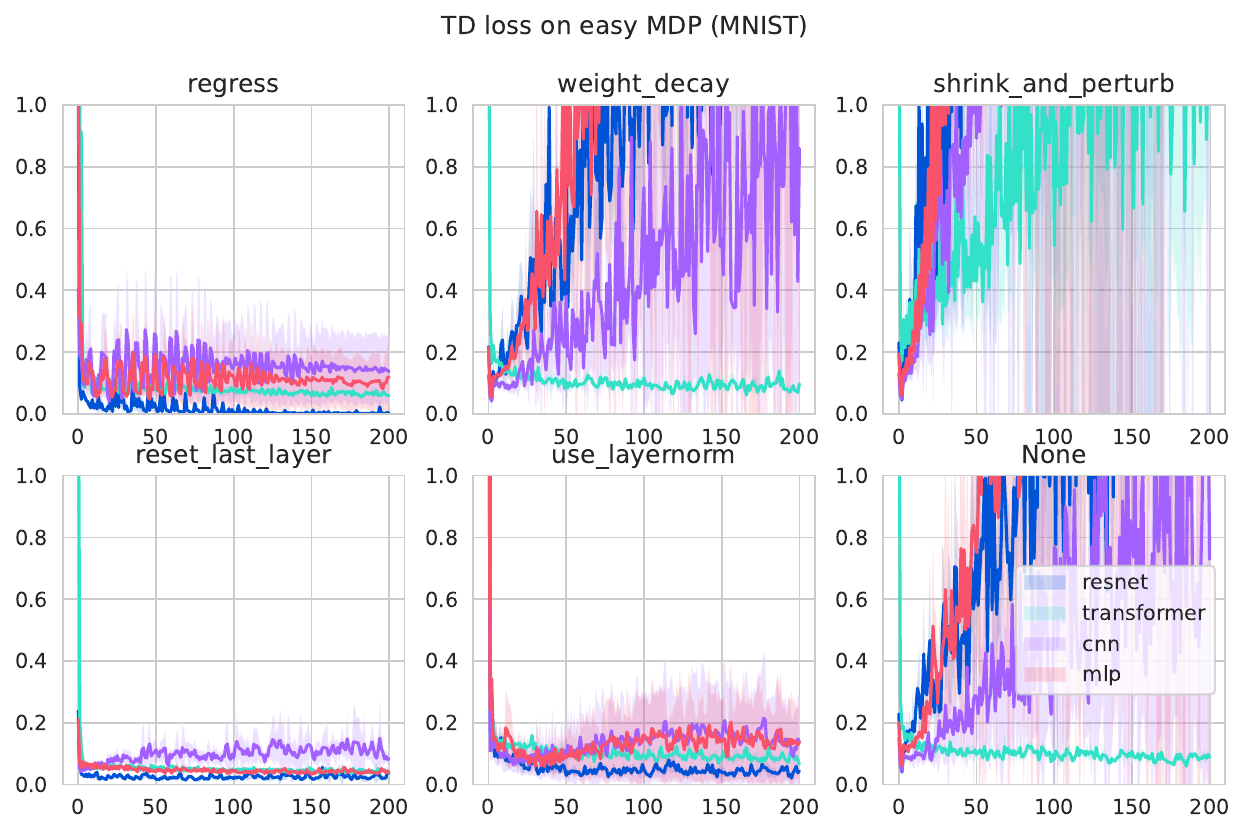}
    \includegraphics[width=0.485\linewidth]{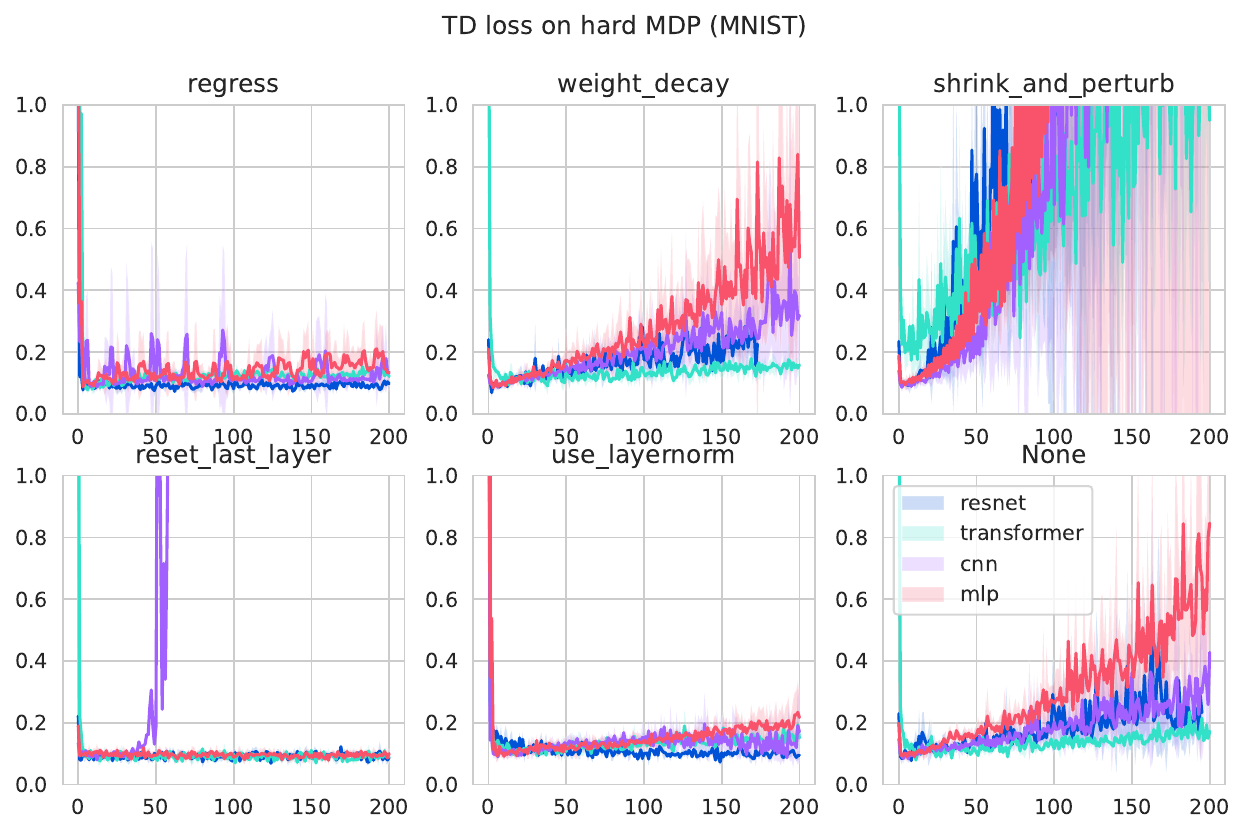}
    \caption{We visualize the TD loss obtained by agents trained on the `easy' and `hard' tasks on MDPs using the MNIST observation space.}
    \label{fig:losses-interventions-mnist}
\end{figure}

\begin{figure}
    \centering
    \includegraphics[width=0.485\linewidth]{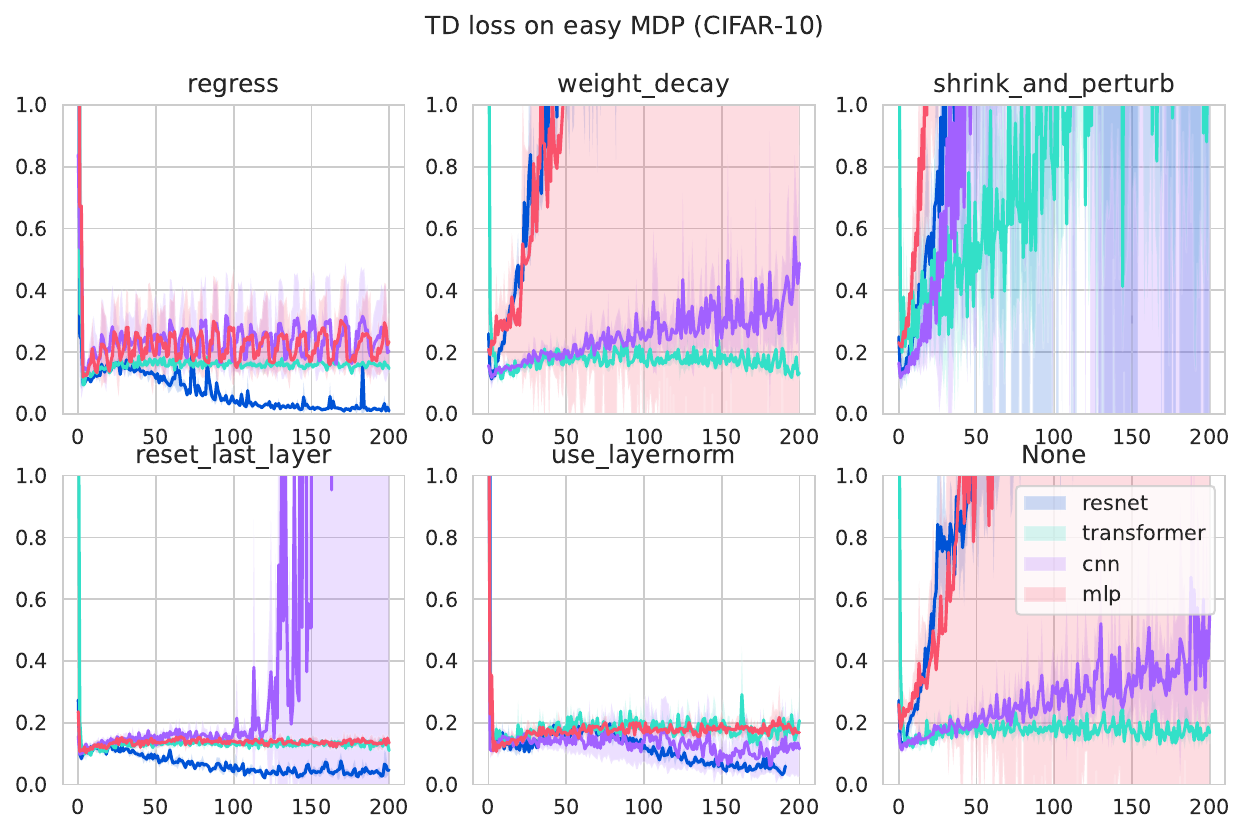}
    \includegraphics[width=0.485\linewidth]{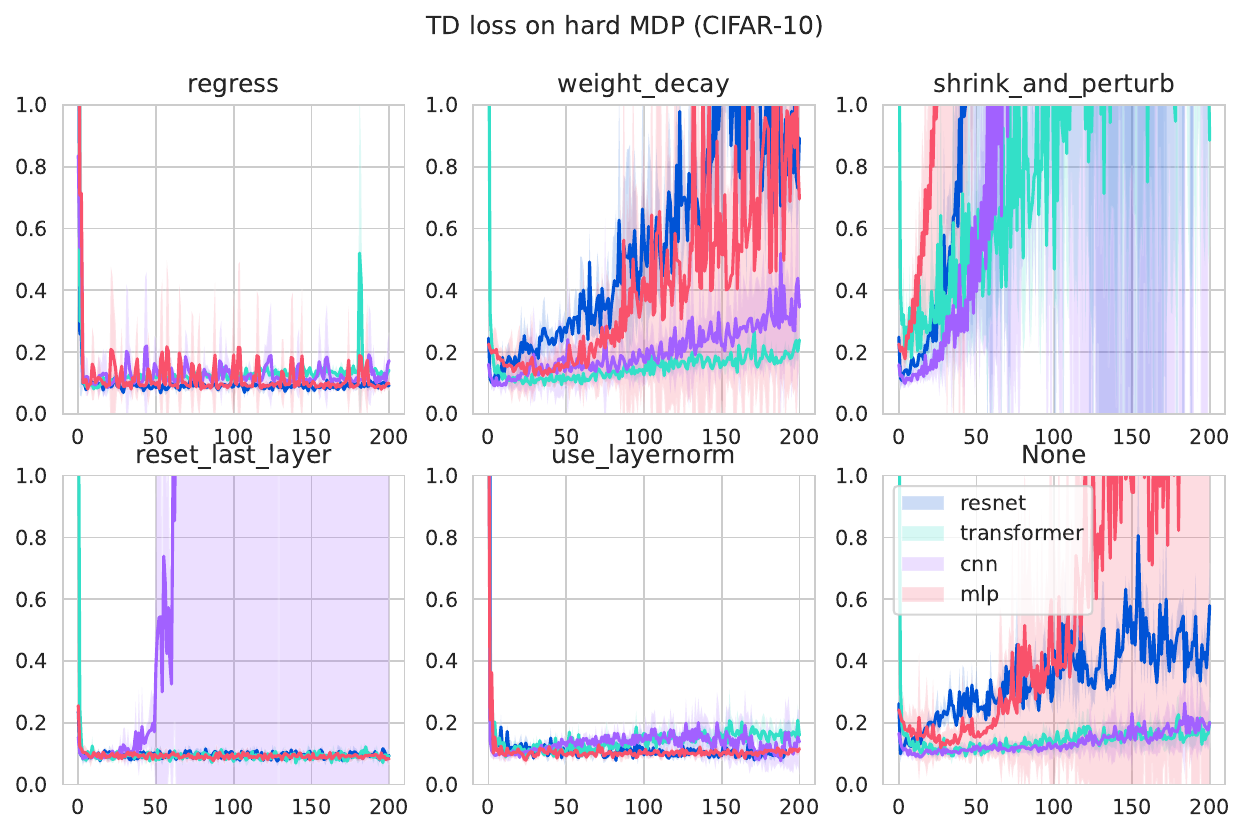}
    \caption{We visualize the TD loss obtained by agents trained on the `easy' and `hard' tasks on MDPs using the CIFAR-10 observation space.}
    \label{fig:losses-interventions-cifar}
\end{figure}

Figure~\ref{fig:losses-interventions-cifar} and ~\ref{fig:losses-interventions-mnist} provides a visualization of the learning curves of different networks during training in the classification MDP experiment. Notably, while performance on the easy and hard reward functions differs dramatically in most agents, the TD losses behave more similarly. This can be attributed to two properties: first, the TD loss changes less in an environment where the agent receives fewer rewards, so the poorer-performing policies induce easier learning problems. Second, the value functions of the two problems are very similar even though the optimal policies are very different. As a result, an `accurate' value function with mean squared error of 0.1 can nonetheless correspond to a policy that does no better than random guessing. The temporal difference losses also give us an indication on whether the agent's predictions are diverging, identifying that layer normalization and the two-hot encodings enable significantly more stable learning, even though this does not always correspond to optimal policies.

\subsubsection{Probe tasks}
\begin{figure}
    \centering
    \includegraphics[width=0.49\linewidth]{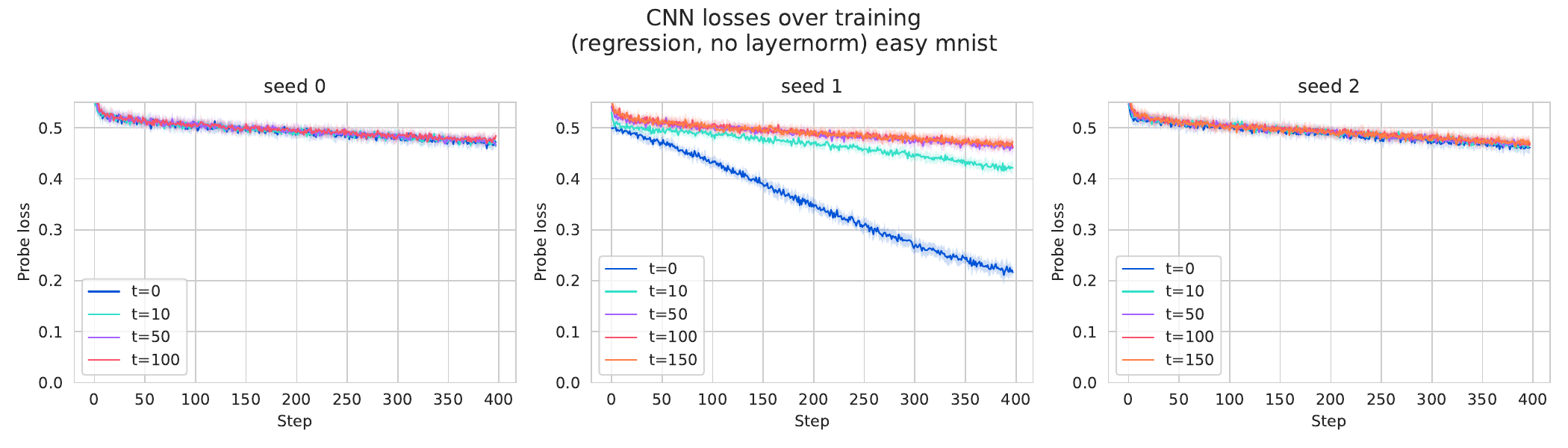}
     \includegraphics[width=0.49\linewidth]{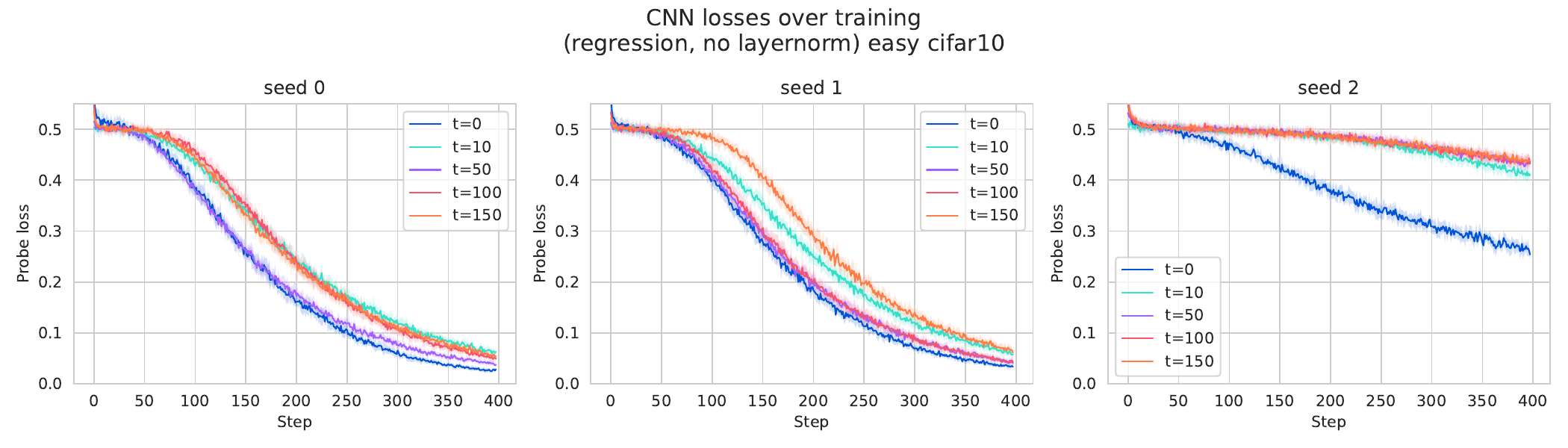}
    \includegraphics[width=0.49\linewidth]{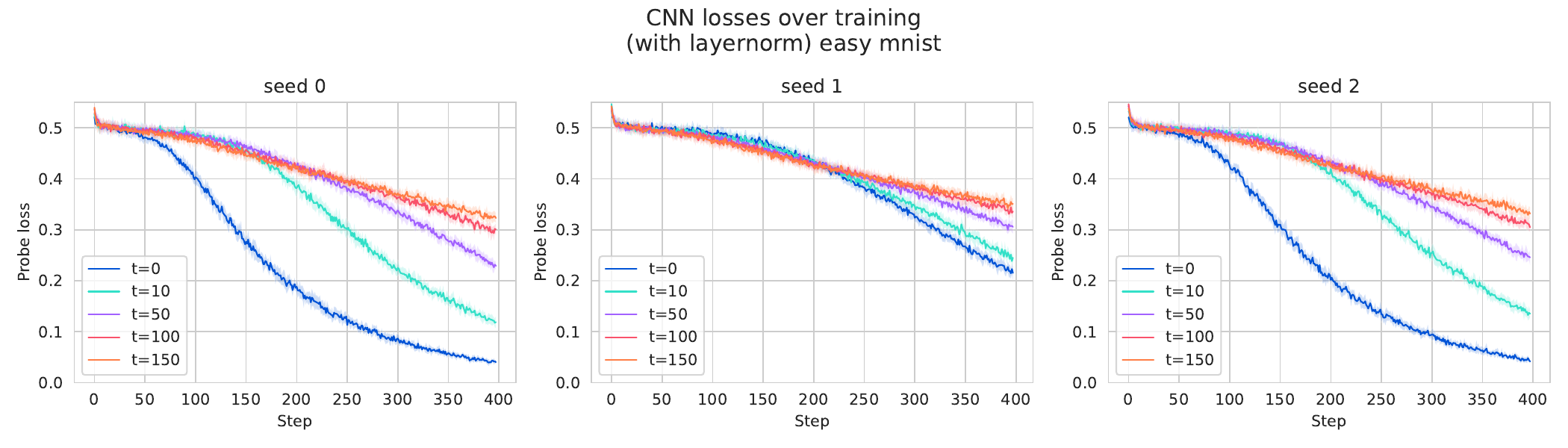}
    \includegraphics[width=0.49\linewidth]{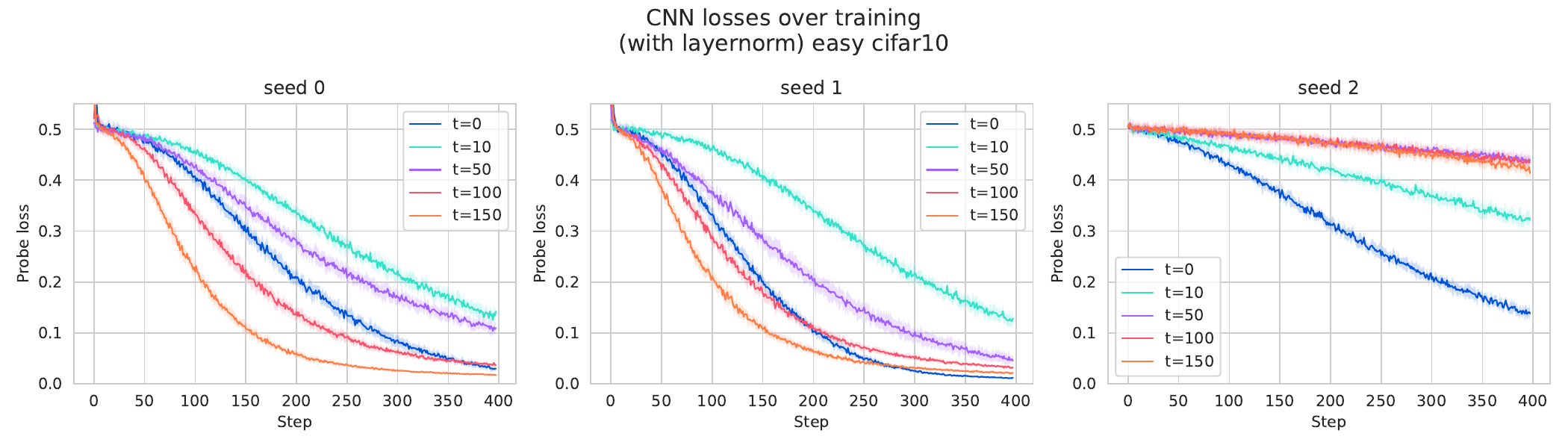}
    \includegraphics[width=0.49\linewidth]{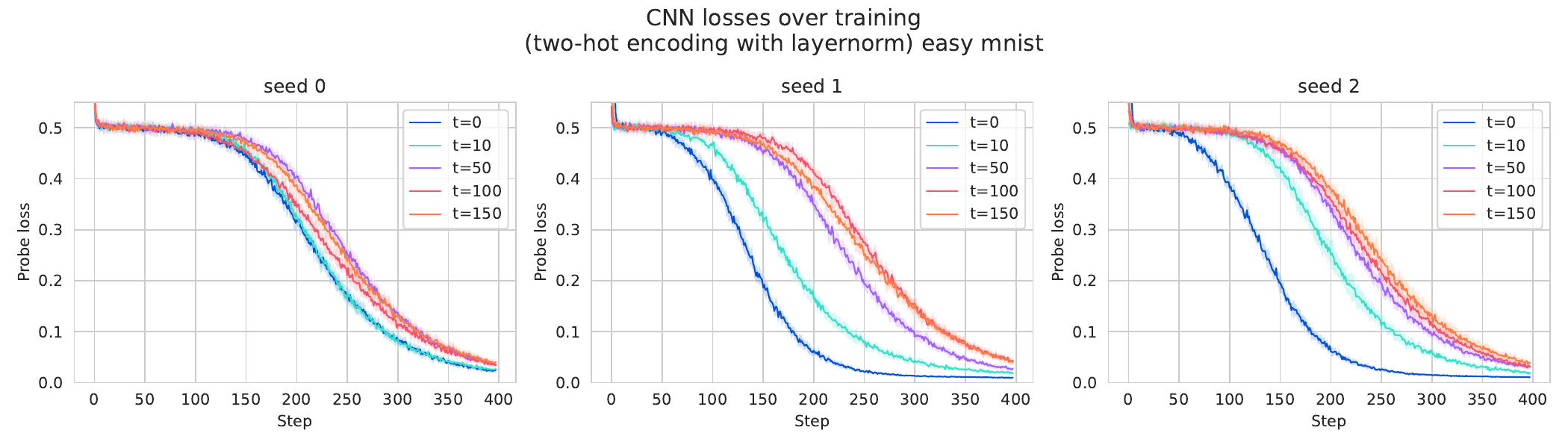}
     \includegraphics[width=0.49\linewidth]{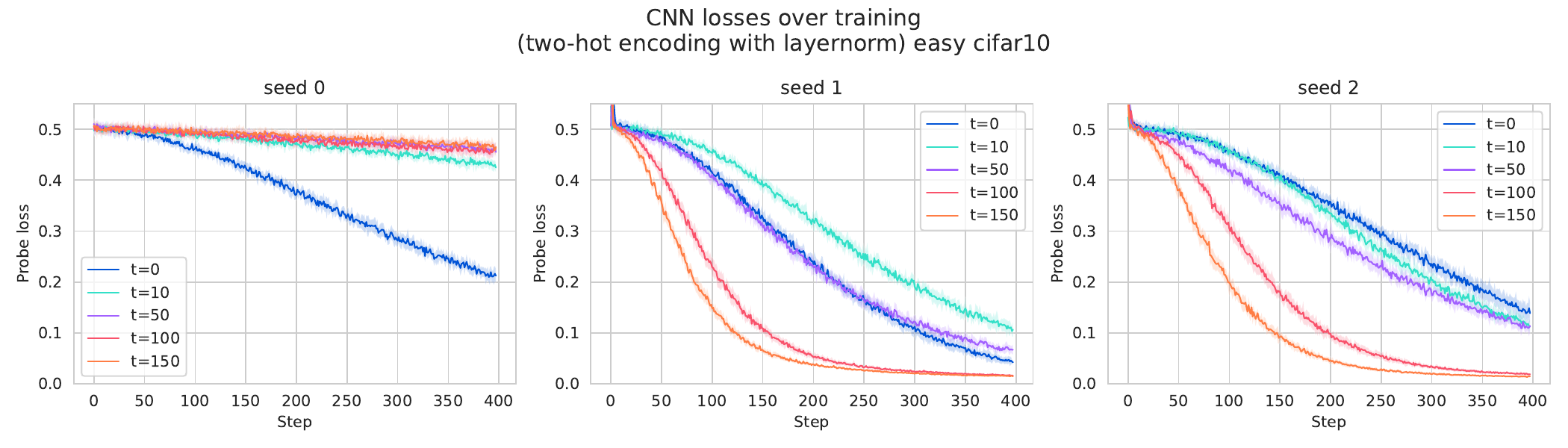}
    \includegraphics[width=0.49\linewidth]{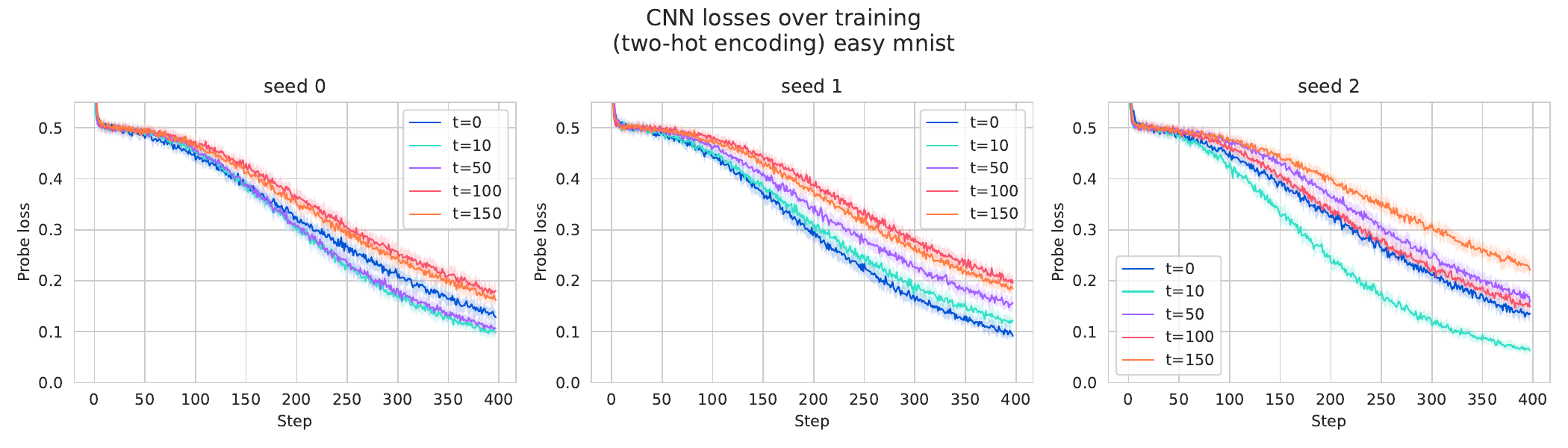}
    \includegraphics[width=0.49\linewidth]{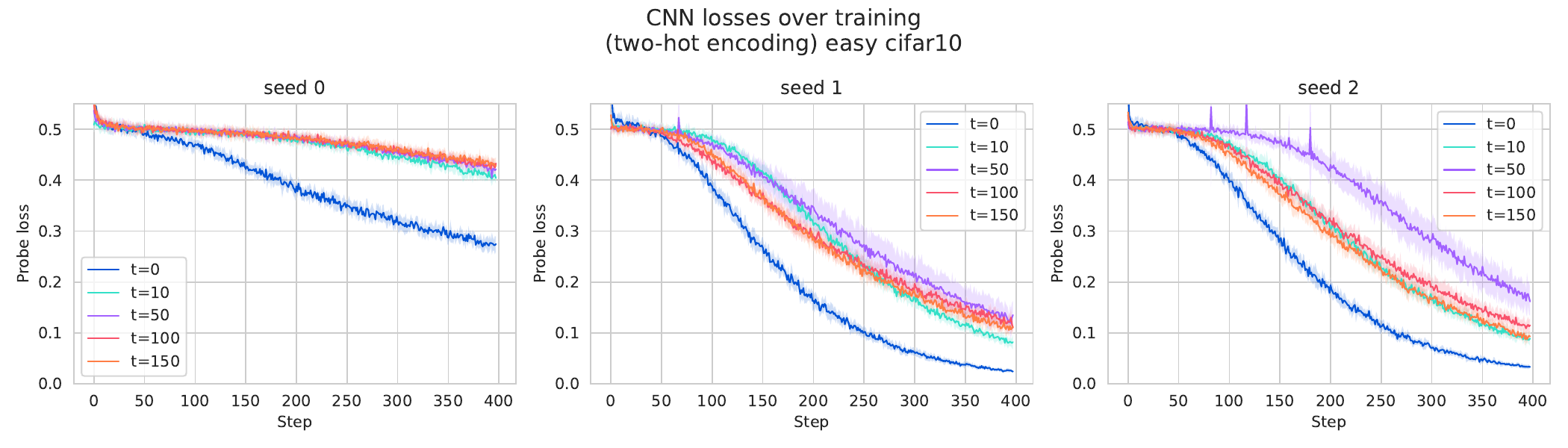}
    \caption{We see markedly different trajectories in networks trained with and without layernorm when tasked with a new optimization objective.}
    \label{fig:cnn-easy}
\end{figure}

\begin{figure}
    \centering
    \includegraphics[width=0.49\linewidth]{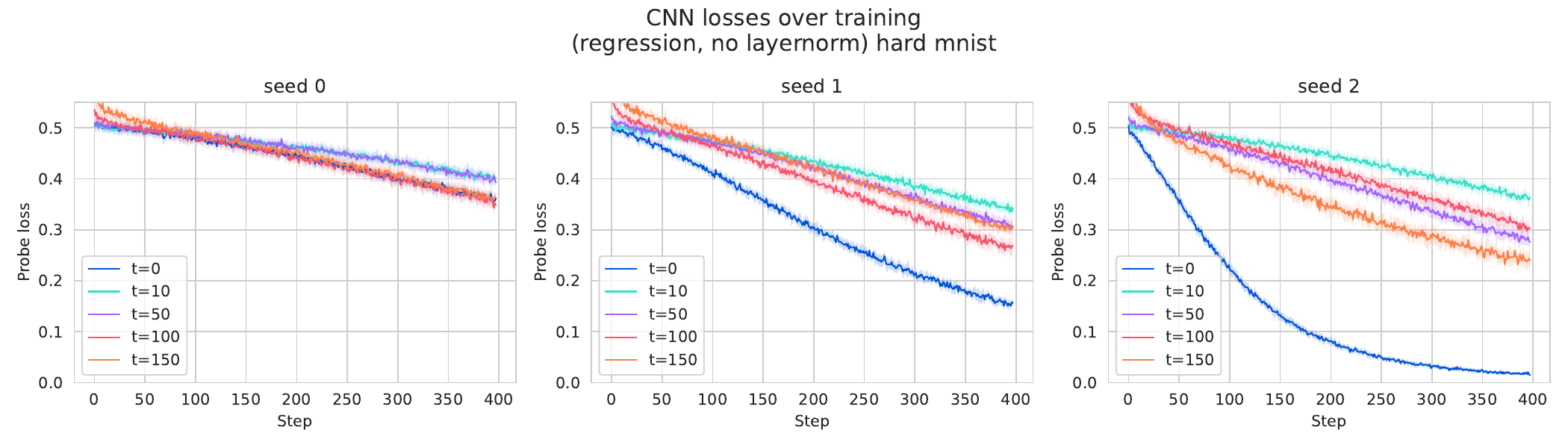}
     \includegraphics[width=0.49\linewidth]{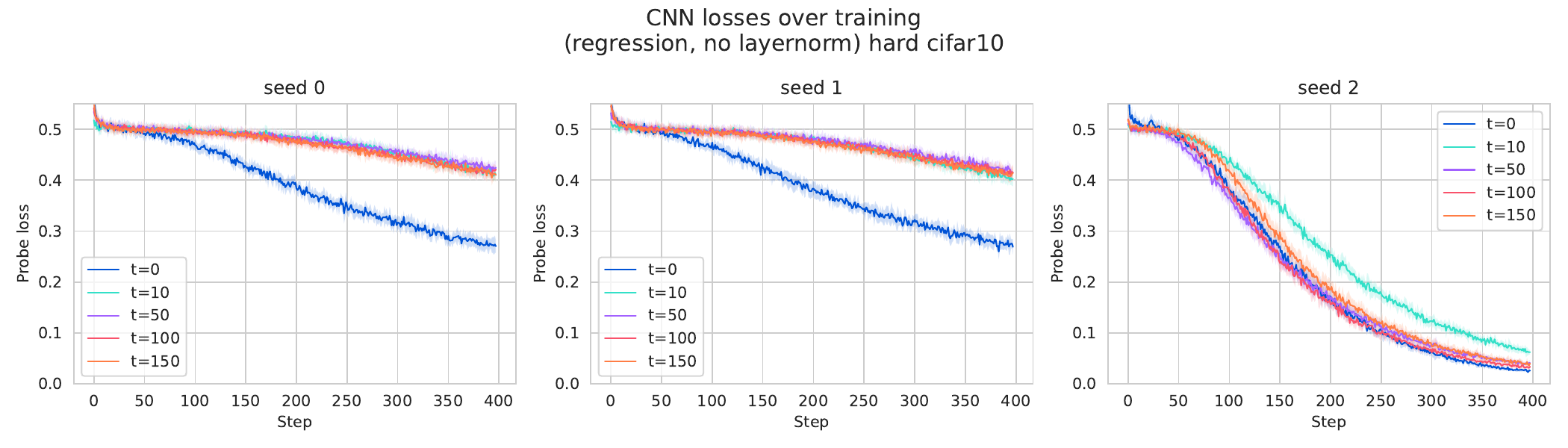}
    \includegraphics[width=0.49\linewidth]{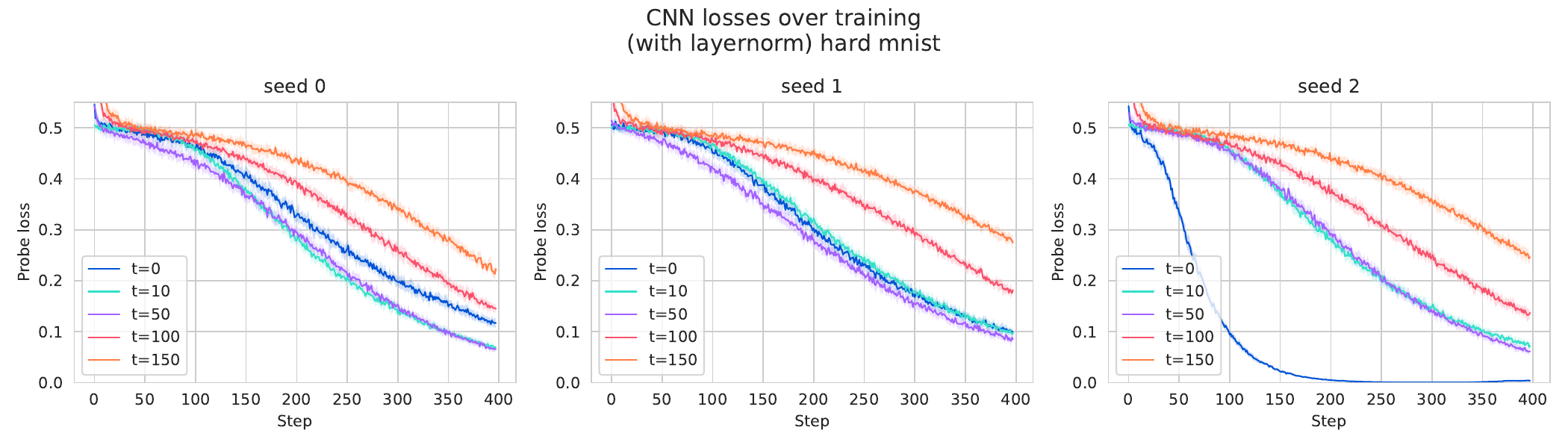}
    \includegraphics[width=0.49\linewidth]{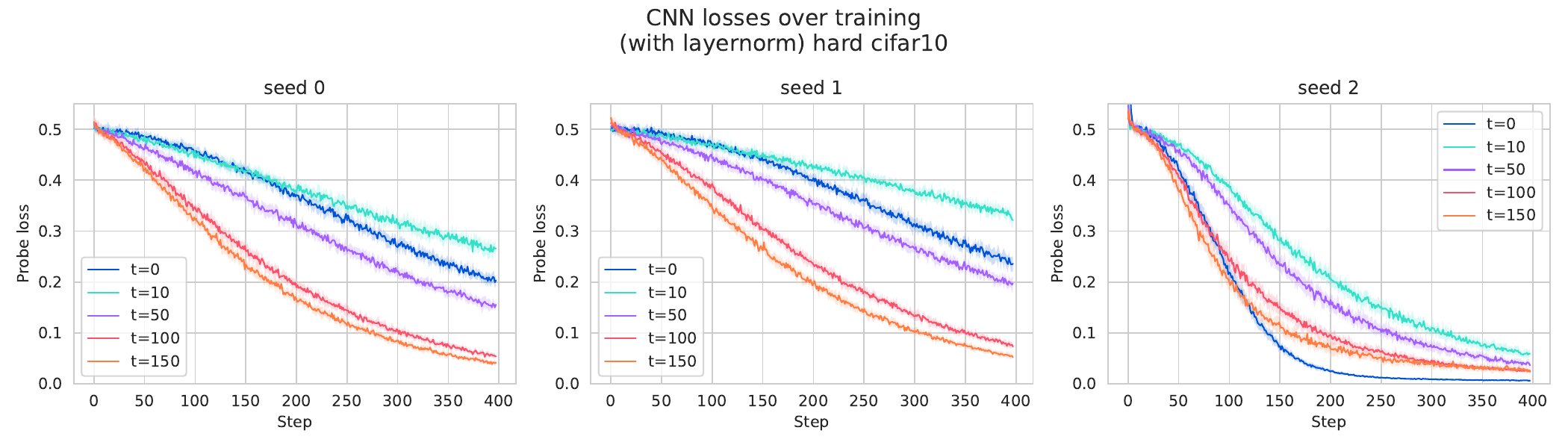}
    \includegraphics[width=0.49\linewidth]{figures/appendix/twohot_layernorm_easy.pdf}
     \includegraphics[width=0.49\linewidth]{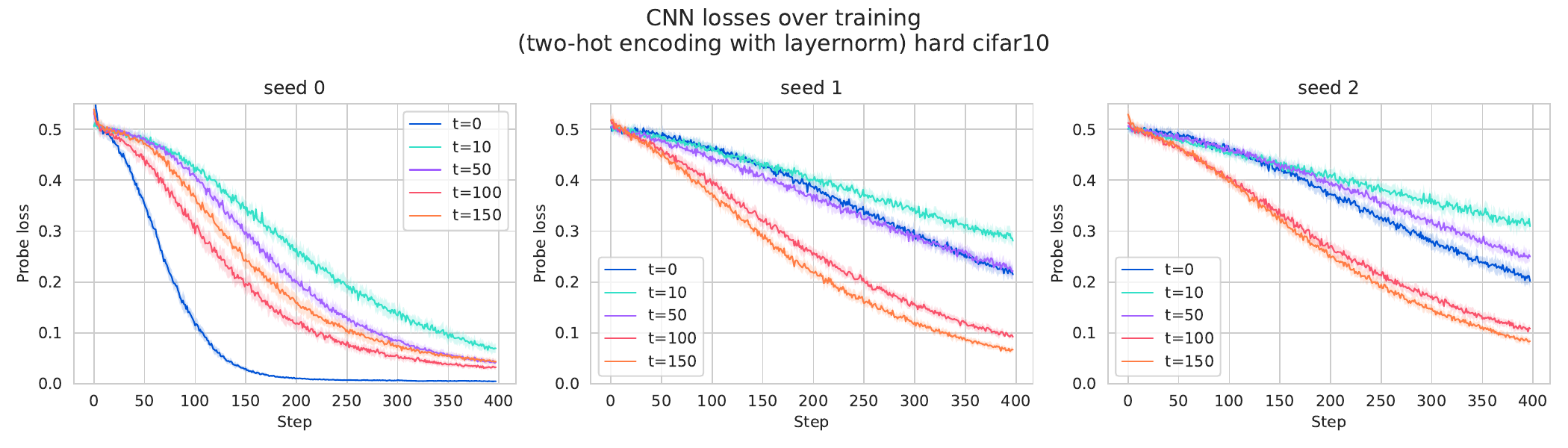}
    \includegraphics[width=0.49\linewidth]{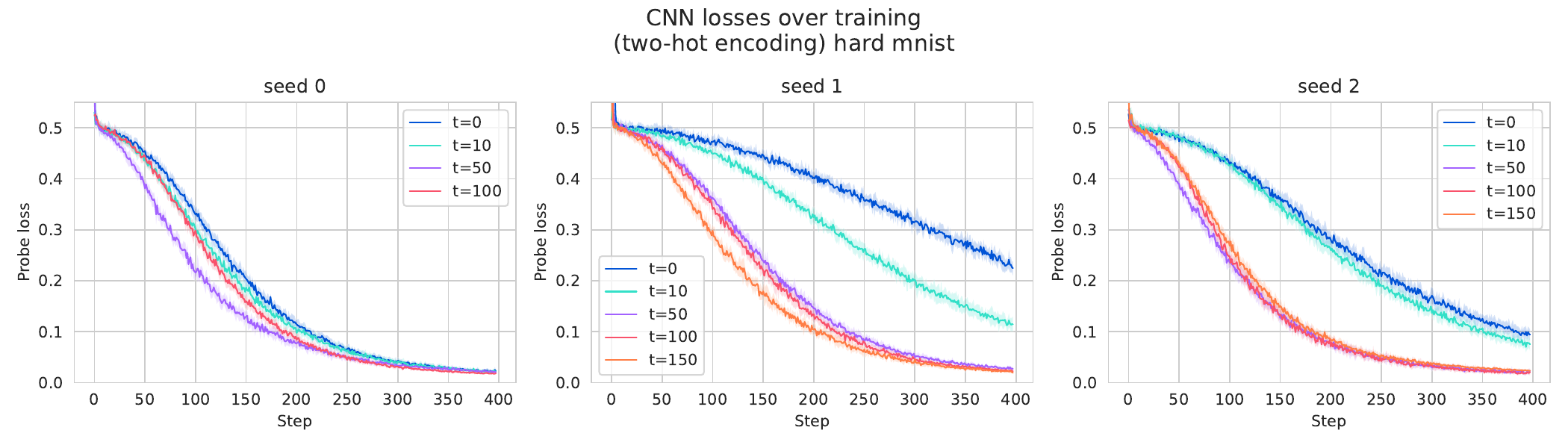}
    \includegraphics[width=0.49\linewidth]{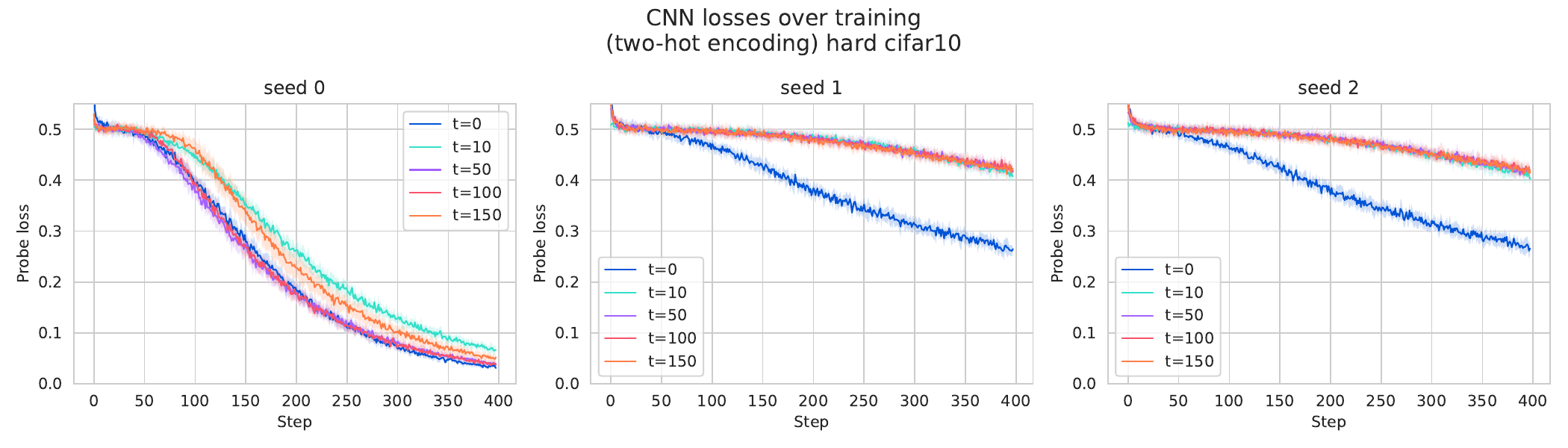}
    \caption{We see markedly different trajectories in networks trained with and without layernorm when tasked with a new optimization objective.}
    \label{fig:cnn-hard}
\end{figure}

We include a visualization of the learning curves of some networks on the probe tasks to illustrate the subtlety of measuring plasticity loss in Figures~\ref{fig:cnn-easy} and~\ref{fig:cnn-hard}. 

\subsection{Qualitative findings in DDQN}
\label{sec:qualitative-dqn}
\begin{figure}
    \centering
    \includegraphics[width=\linewidth]{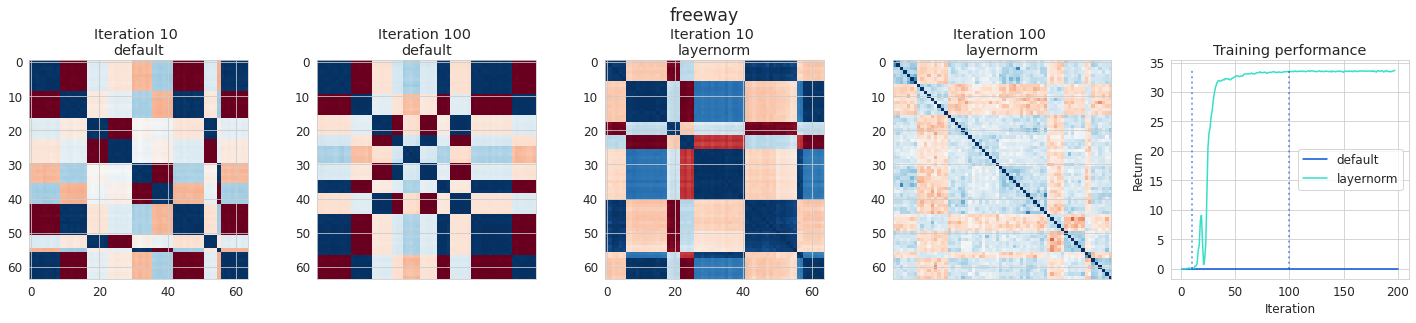}
    \includegraphics[width=\linewidth]{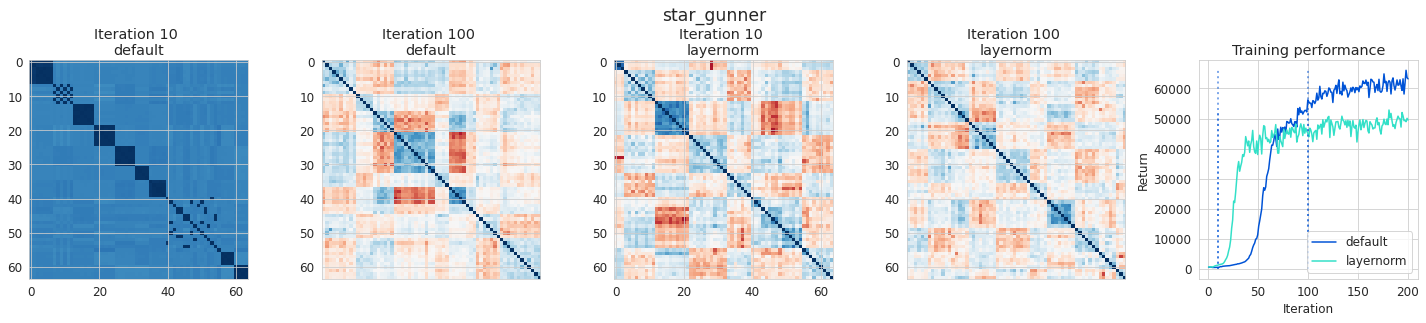}
    \includegraphics[width=\linewidth]{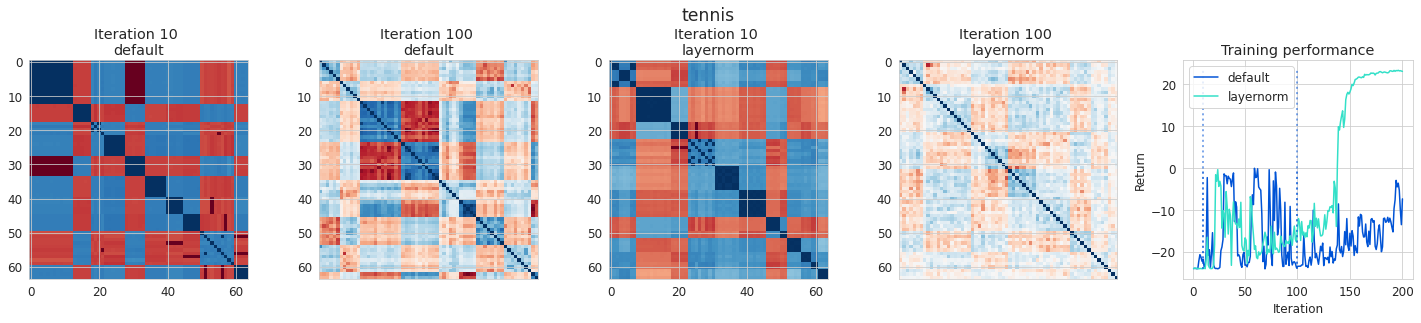}
    \caption{Gradient covariance plots vs performance for three sample games from the Arcade Learning Environment, which highlight the role of the gradient structure in learning progress. We find that when agents fail to learn, they tend to exhibit highly degenerate gradient structure, corresponding to the large off-diagonal values in the heatmaps visualized here.}
    \label{fig:heatmaps-2}
\end{figure}

\begin{figure}
    \centering

    \includegraphics[width=\linewidth]{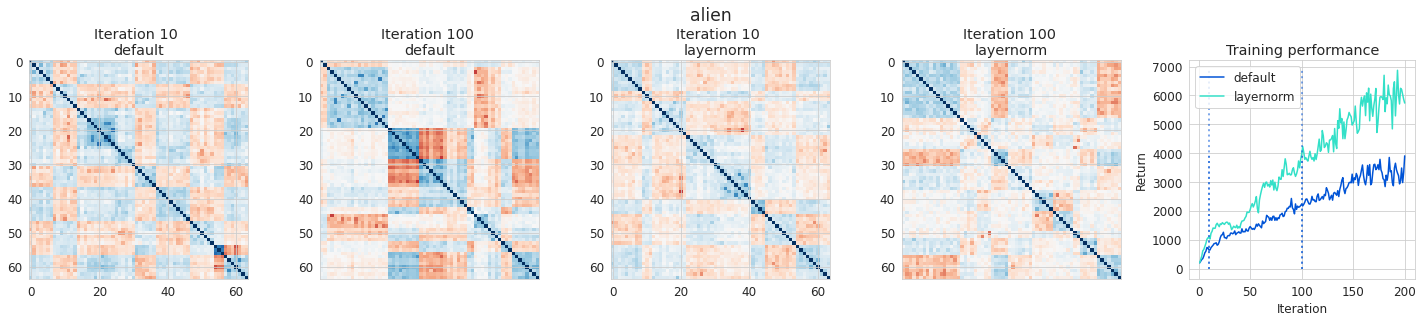}
    \includegraphics[width=\linewidth]{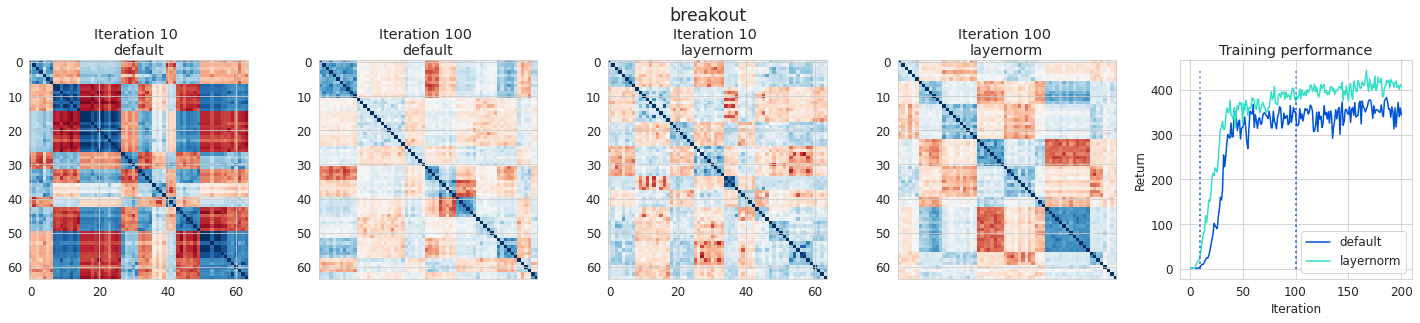}
    \caption{Gradient covariance plots vs performance for three sample games from the Arcade Learning Environment, which highlight the role of the gradient structure in learning progress. In games where agents make consistent positive progress increasing their returns, we see covariance structures with heavier weight on the diagonal.}
    \label{fig:heatmaps}
\end{figure}

\begin{figure}
    \centering
    \includegraphics[width=0.89\linewidth]{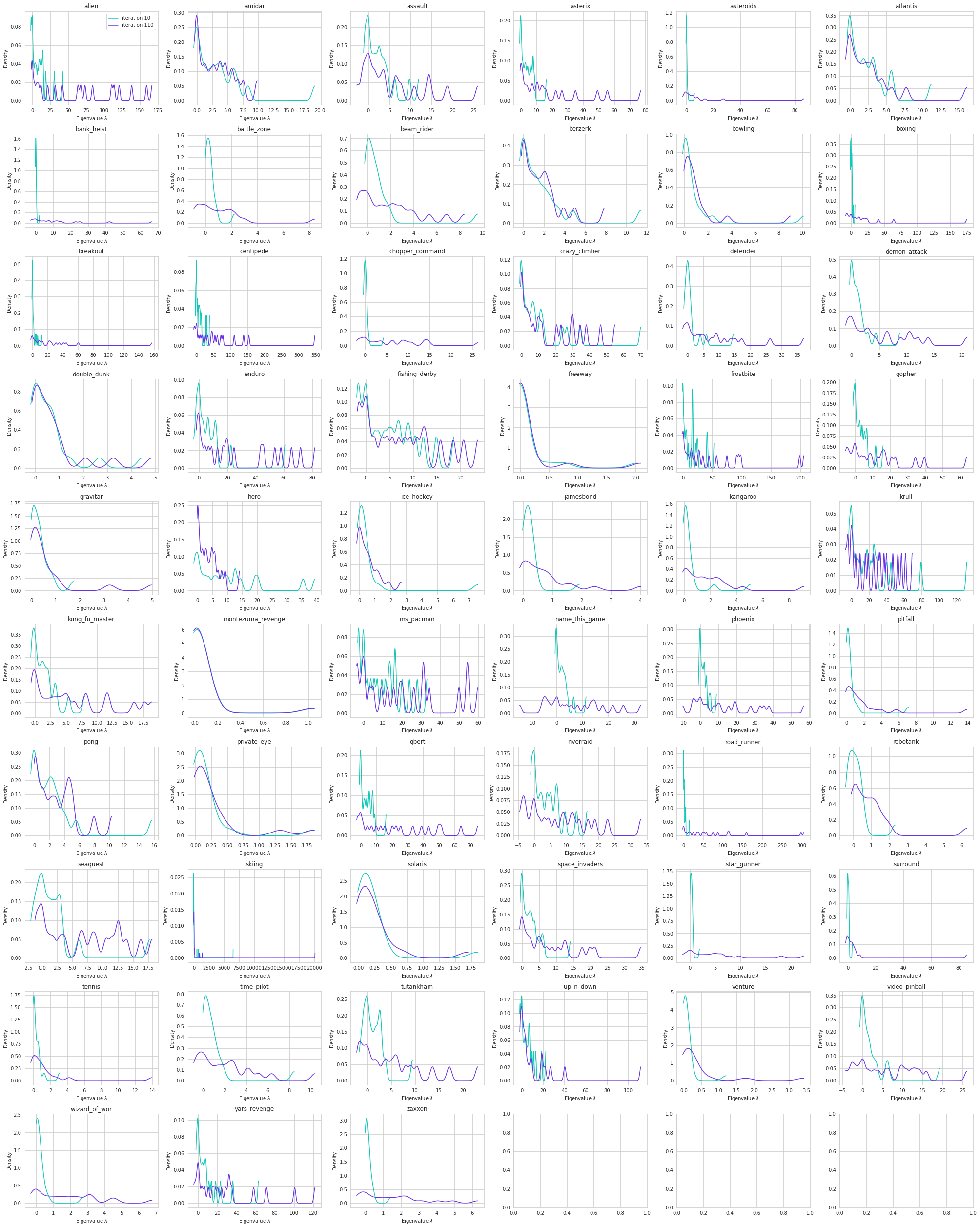}
    \caption{Visualization of Hessian approximations for a double DQN agent \textbf{without} layer normalization.}
    \label{fig:hessian-ddqn-default}
\end{figure}

\begin{figure}
    \centering
    \includegraphics[width=0.89\linewidth]{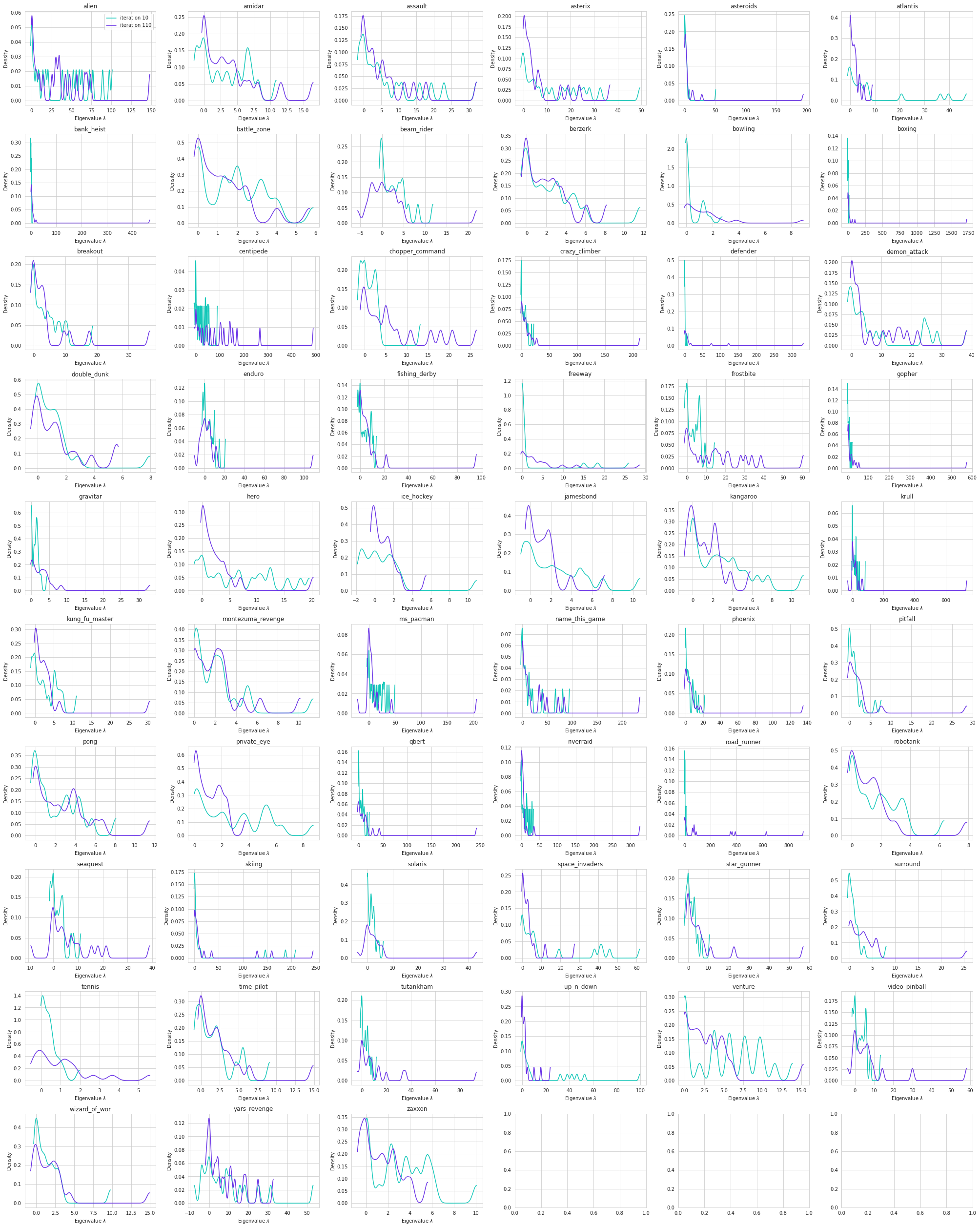}
    \caption{Visualization of Hessian approximations for a double DQN agent with layer normalization.}
    \label{fig:hessian-ddqn-layernorm}
\end{figure}

In addition to the gradient covariance analysis presented in the main body, we show more detailed learning curves and illustrate the evolution of the gradient covariance over time in Figures~\ref{fig:heatmaps} and~\ref{fig:heatmaps-2}. We also visualize the spectrum of the network Hessian in Figures~\ref{fig:hessian-ddqn-default} and~\ref{fig:hessian-ddqn-layernorm}. We observe particularly intriguing trends in the gradient covariance heat maps shown in Figure~\ref{fig:heatmaps-2}. We note that the covariance structure of gradients varies significantly across environments, networks, and even random seeds. In many situations, gradients appear to be largely colinear, corresponding to significant interference (both positive and negative) between transitions in a minibatch. 

One phenomenon we noticed in several environments was a tendency for network gradients to start off highly colinear, and then to become more independent later in training. In general, networks which stay in this colinear phase longer are also those whose learning curves struggle to `take off'. Notably, in the game Freeway which is known to produce extremely high variance outcomes hwerein agents either maximize the game score or fail to learn at all, we saw a one-to-one mapping between gradient degeneracy and learning progress. All random seeds where the agent made learning progress exhibited heavier weight on as opposed to off the diagonal, whereas the random seeds which did not ever improve preserved their initial degenerate gradient structure. A representative example can be observed in the first row of Figure~\ref{fig:heatmaps-2}. In general, networks with layer normalization exhibited a slight bias towards less degenerate gradients. However, it is not clear whether gradient degeneracy is a symptom or a cause of performance plateaus. Further investigation into this phenomenon presents an exciting avenue for future work.

We did not observe any obvious correlations between the spectrum of the Hessian and the agent's performance, but include the computed spectra for a single seed of each game in Figures~\ref{fig:hessian-ddqn-default} and \ref{fig:hessian-ddqn-layernorm}. 

 \end{document}